\pdfoutput=1
\documentclass[sigconf]{acmart}
\usepackage{hyperref}
\usepackage{makecell}
\usepackage{multirow}
\usepackage{algorithm}
\usepackage{algorithmicx}
\usepackage{color}
\usepackage{inputenc}
\usepackage{amsmath}
\usepackage{subfig}
\usepackage{amsmath,bm}

\AtBeginDocument{%
  \providecommand\BibTeX{{%
    \normalfont B\kern-0.5em{\scshape i\kern-0.25em b}\kern-0.8em\TeX}}}

\begin{document}

\title{Low-Resource Multi-Granularity Academic Function Recognition Based on Multiple Prompt Knowledge}


\author{Jiawei Liu, Zi Xiong}
\authornote{The first two authors contributed equally to this research.}

\author{Yi Jiang, Yongqiang Ma, Wei Lu, Yong Huang}

\author{Qikai Cheng}
\authornote{Corresponding author.}

\affiliation{%
  Wuhan University, School of Information Management, China\\
  {\{laujames2017,zixiong,yijiang,weilu,yonghuang1991,chengqikai\}@whu.edu.cn, mayq97@qq.com}\\
    \country{}
}

\renewcommand{\shortauthors}{Liu, et al.}

\begin{abstract}
[Purpose] Fine-tuning pre-trained language models (PLMs), e.g., SciBERT, generally requires large numbers of annotated data to achieve state-of-the-art performance on a range of NLP tasks in the scientific domain. However, obtaining the fine-tune data for scientific NLP tasks is still challenging and expensive.
Inspired by recent advancements in prompt learning, in this paper, we propose the Mix Prompt Tuning (MPT), which is a semi-supervised method to alleviate the dependence on annotated data and improve the performance of multi-granularity academic function recognition tasks with a small number of labeled examples.
[Method] Specifically, the proposed method provides multi-perspective representations by combining manual prompt templates with automatically learned continuous prompt templates to help the given academic function recognition task take full advantage of knowledge in PLMs. Based on these prompt templates and the fine-tuned PLM, a large number of pseudo labels are assigned to the unlabeled examples. Finally, we fine-tune the PLM using the pseudo training set. We evaluate our method on three academic function recognition tasks of different granularity including the citation function, the abstract sentence function, and the keyword function, with datasets from the computer science domain and the biomedical domain.

[Findings] Extensive experiments demonstrate the effectiveness of our method and statistically significant improvements against strong baselines. 
In particular, it achieves an average increase of 5\% in Macro-F1 score compared with fine-tuning, and 6\% in Macro-F1 score compared with other semi-supervised methods under low-resource settings.

[Originality/value] In addition, MPT is a general method that can be easily applied to other low-resource scientific classification tasks.\footnote{This article has been accepted by The Electronic Library and the full article is now available on Emerald Insight.}

\end{abstract}



\maketitle

\section{Introduction}
With the exponential expansion of the research community and the volume of scientific publications, it becomes harder and harder to acquire knowledge timely and accurately from scientific literature. In responding to the growing problem of information overload, research and development of efficient strategies \citep{beltagy2019scibert,lu2018functional} and intelligent tools \citep{lahav2021search,yin2021mrt} to accelerate scientific breakthroughs have attracted increasing attention from industry and academia. Behind these strategies and tools, there are various fundamental scientific NLP tasks and datasets support. Identifying the multi-granularity function of a keyword \citep{wei2020recognition}, a sentence \citep{jin2018hierarchical}, or a citation \citep{cohan2019structural} in the scientific paper is critical for downstream tasks, such as impact prediction \citep{huang2022disclosing,qin2022structure,zhou2020evaluating}, novelty measurement \citep{luo2022combination} and emerging topic prediction \citep{liang2021combining,huo2022hotness}.

\begin{figure*}[!t]
\centering
\includegraphics[width=6in]{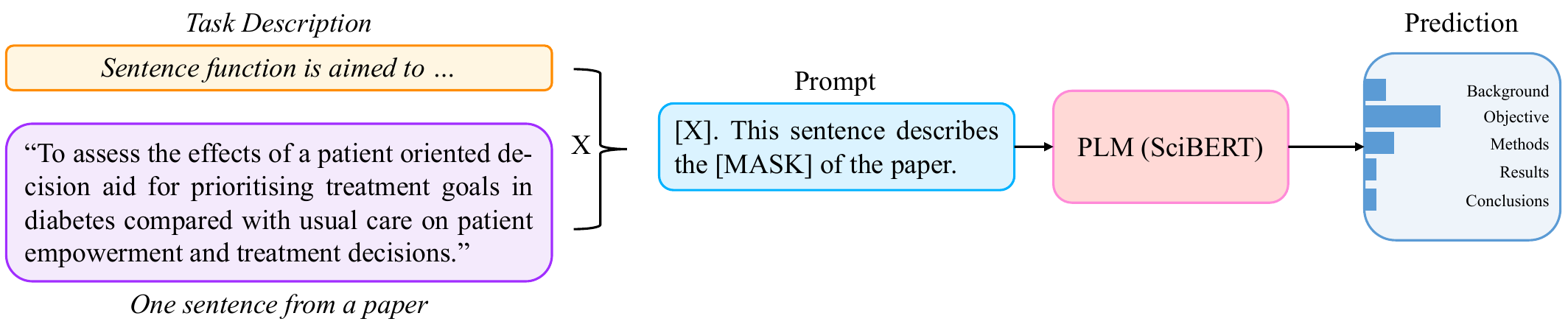}
\caption{An example of prompt learning for sentence function recognition (a typical scientific text classification task). $X$ denotes the concatenation of the task description and the sentence from a paper. We utilize a pre-trained language model to predict the \textit{MASK} token in the prompt template. Note that the candidate mask tokens, i.e., pre-defined verbalizer, are restricted in the set \{\textit{Background, Objective, Methods, Results, Conclusions}\}.}
\label{fig_exemplar}
\end{figure*}

Deep neural networks are adopted to achieve great progress in these scientific NLP tasks, but training such models requires a large mount of annotated data. In the past few years, pre-trained language models (PLMs), such as GPT \citep{radford2018improving} and BERT \citep{devlin2019bert}, self-supervised trained on large-scale corpora have emerged as a powerful instrument for language understanding. It is because that PLMs can capture different levels of syntactic \citep{hewitt2019structural}, linguistic \citep{jawahar2019does}, and semantic \citep{yenicelik2020does} from Large-scale corpus. As a result, PLM fine-tuning has shown awesome performance on almost all important NLP tasks and becomes a common way of the NLP community instead of training models from scratch \cite{qiu2020pre}.
Especially in scientific domain, fine-tuning these PLMs, e.g., SciBERT \citep{beltagy2019scibert} and BioBERT\citep{lee2020biobert}, with additional task-specific data achieves new state-of-the-art (SOTA) performances. However, obtaining the fine-tune data for scientific NLP task is still challenging and expensive. It is common in real-world scenarios that there is no annotated fine-tuning data or only a small number of annotated examples. 
Thus, it is necessary to explore method suitable for low-resource or few-shot scientific NLP tasks.

To address this problem, recently, scholars \citep{petroni2019language,radford2019language,brown2020language} propose to bridge the gap of objective forms in pre-training and fine-tuning, and make full use of PLMs by reformulating tasks as fill-in-the-blanks problems, i.e., prompt learning. In this way, downstream tasks look more like those solved tasks during the original language model (LM) or mask LM training with the help of a prompt \citep{liu2021pre}. For instance, as Figure \ref{fig_exemplar} shown, when recognizing the function of a sentence from a paper, ``To assess the effects of a patient oriented decision aid for prioritising treatment goals in diabetes compared with usual care on patient empowerment and treatment decisions'', referring to task descriptions, we could continue with a prompt ``This sentence describes the [\textit{MASK}] of the paper.'', and use the pre-trained mask LM to fill the [\textit{MASK}] blank with a function word, e.g., ``objectives''. A series of research works based on manual prompts have achieved promising performance on few-shot sentiment classification \citep{schick2020automatically}, fake news detection \citep{jiang2022fake}, and natural language inference \citep{han2021ptr}. To ease the manual effort of suitable prompt design, some works propose to search prompt based gradient search \citep{shin2020autoprompt,liu2021gpt,hambardzumyan2021warp}. Furthermore, \citet{schick2021exploiting} introduce a semi-supervised framework utilizing natural language prompt to annotate unlabeled data. Their method substantially outperforms unsupervised, supervised, and strong semi-supervised baselines, e.g., UDA \citep{xie2020unsupervised} and MixText \citep{chen2020mixtext}. However, scientific NLP tasks, such as keyword function recognition and citation function recognition, in low-resource settings are under-explored. Moreover, it could take more effort to find the most appropriate prompt to allow the PLM to solve the scientific NLP tasks than general NLP tasks.

In this paper, we propose a semi-supervised method, named \textit{Mix Prompt Tuning} (MPT), to alleviate the dependence on annotated data and improve the performance of scientific NLP tasks in low-resource settings. In addition to be able to combine the expert knowledge to design the most appropriate manual prompt template of academic function, our proposed method also adopts automatically learned \textit{soft} and manually designed \textit{hard} prompt templates.
The \textit{hard} templates are manually crafted and make use of the expert knowledge to help us understand what the task is about.
The \textit{soft} templates are meaningless to humans but could be informative to the PLM. These prompt templates provide multi-perspective representations to help the given academic function recognition task take full advantage of knowledge in the PLM. Following the semi-supervised training procedure of iterative pattern-exploiting training (iPET) \citep{schick2021exploiting}, based on various templates, we first fine-tune a separate PLM for each template on a small training set. Second, fine-tuned models are randomly sampled to assign pseudo soft labels to a certain number of unlabeled data. By this means, the original training dataset can be enlarged to train new generation of fine-tuned models. This step is repeated for several times to make all models learn from each other. Finally, the ultimate enlarged dataset with pseudo labels annotated by the last generation models is used to train a standard classifier in knowledge distillation manner. Evaluation is conducted with the standard classifier on the corresponding test set. Our proposed MPT, utilizing multiple prompt templates to annotate unlabeled data, is similar to iPET \citep{schick2021exploiting}. Differently, compared with this method that only adopts the manual prompt, we combine the manual prompt with automatically learned continuous prompt, which can provide multi-perspective representations and take full advantage of knowledge in the PLM and unlabeled data.

To the best of our knowledge, this is the first study to introduce prompt learning with mixing multiple types of prompt templates for scientific classification tasks in a more practical scenario, i.e., low-resource or few-shot settings. We conduct extensive experiments on a suite of different granularity academic function recognition tasks, including word function recognition, structure function recognition, and citation function recognition, to demonstrate the effectiveness of our method and statistically significant improvements against strong unsupervised, supervised, and semi-supervised baselines.

To summarize, our contributions are mainly as follows:
\begin{enumerate}
    \item[(1)] We propose a semi-supervised solution to alleviate the dependence on annotated data and improve the performance of scientific classification tasks in low-resource settings.
    \item[(2)] We combine the manual prompt with automatically learned continuous prompt to provide multi-perspective representations and take full advantage of knowledge in the PLM.
    \item[(3)] We perform extensive experiments on a suite of different granularity academic function recognition tasks to demonstrate the effectiveness of our method and statistically significant improvements against strong baselines.
\end{enumerate}

This article is organized as follows: the related work presents an brief literature review; the method section describes the semi-supervised hybrid prompt learning method for academic function recognition; the experiments section describes the tasks, datasets, experimental settings; the results and analysis section provide insight of the experimental results; the conclusion section concludes this work and points out the direction of future work.

\section{Related work}

\subsection{Academic function recognition}
In this paper, we mainly focus on three levels of academic function recognition tasks, which are the citation function recognition, the sentence function recognition and the keyword function recognition.

\textbf{Citation function}: 
Numerous previous studies have introduced citation classification schemes \citep{teufel2006automatic,jurgens2016citation,cohan2019structural,pride2020authoritative} as a way to identify the meaning or purpose of a specific citation. Note that for terminological consistency, we refer to the ``citation classification'' and the ``citation intent classification'' as the ``citation function recognition''. \citet{jurgens2016citation} and \citet{cohan2019structural} adopt ML based and DL based methods to recognition the citation function, respectively. Furthermore, \citet{yu2020identifying} propose a interactive hierarchical attention model with aggregating the heterogeneous contexts to recognize the intention of citing behaviors and retweeting behaviors. Similarly, \citet{beltagy2019scibert} also fine-tune the SciBERT to recognize the citation function achieve SOTA performance.

\textbf{Structure function}: Structure function recognition tasks for scientific publications mainly concentrate on the abstract sentence function recognition and body sentence structural function recognition. For abstract sentence function, previous studies are mainly based on machine learning (ML) methods \citep{ruch2007using,hassanzadeh2014identifying,liu2013abstract} and deep learning (DL) methods \citep{lui2012feature, dernoncourt2016neural}. Recently, SciBERT \citep{beltagy2019scibert} is also adopted to directly classify the sentence function and achieve competitive performance. In terms of body sentence structural function recognition, there are also a series of works based on ML \citep{huang2016sec,huang2016para,lu2018functional} and DL \citep{qin2020using,qian2020structure} based methods.
Moreover, scholars also consider the context semantics and structural relation of surrounding sentences \citep{jin2018hierarchical,jiamin2019research} to boost the sentence function recognition performance.

\textbf{Keyword function}: \citet{kondo2009technical} were first to conduct research on automatic recognition of lexical  semantic functions. They utilize conditional random filed (CRF) to divide the words in the academic title into four labels of ``domain'', ``problem'', ``method'', and ``others''. \citet{nanba2010automatic} adopt support vector machine (SVM) to recognize the ``technology'' and ``effect'' word in academic and patent literature. \citet{cheng2021recognition} use recurrent neural network (RNN) based sequence-to-sequence model to obtain the problem and method words in academic texts.
\citet{wei2020recognition} and \citet{zhang2021multiword} utilize BERT-based model to classify the problem and method semantic function carried by the keywords in academic literature.

However, above mentioned ML, DL, and even PLMs based methods require a large number of annotated data to achieve competitive performance.

\subsection{Low-Resource text classification}
Since obtaining the training datasets for these tasks are challenging and expensive, it is important to develop systems that perform decent in low-resource settings, where few labeled examples are available. Intuitively, we can adopt NLP techniques to increase the amount of training data, i.e., data augmentation, including synonym replacement, random insertion, random swap, random deletion \citep{wei2019eda}, paraphrasing formulation, and back translation \citep{chen2020semi,chen2020mixtext,xie2020unsupervised}.
Another typical way, widely adopted in computer vision and general NLP tasks, is to design meta-learning paradigm that can learn and adapt to new environments rapidly with a few training examples  \citep{bao2019few,yao2021knowledge,sun2021meda,zhang2022contrastive}. These methods usually train a meta-learner that extracts knowledge from various related sub-tasks during meta-training and leverages the knowledge to learn new tasks during meta-testing quickly. Considering that the function of academic publication is already relatively well-defined, there is no need to identify new function category. In this paper, we mainly adopt the data augmentation based methods as target baselines.

\subsection{Prompt learning}
Fine-tuning the PLMs is a conventional approach to leverage the rich knowledge during pre-training and has achieved satisfying results on supervised tasks \citep{devlin2019bert,han2021pre}. However, tuning the extra classifier requires adequate training examples to achieve decent performance, it is still challenging to apply fine-tuning in low-resource settings, including few-shot and zero-shot learning scenarios \citep{brown2020language,yin2019benchmarking}. Recently, a series of studies \citep{petroni2019language,schick2021exploiting,liu2021gpt,jiang2022fake} using prompts to bridge the gap of objective forms in pre-training and fine-tuning and make full use of PLMs. Manual prompts have achieved promising performance in low-resource sentiment classification and natural language inference \citep{schick2020automatically,han2021ptr}. A typical prompt consists of two parts: a template and a set of label words, i.e., verbalizer. \citet{zhou2022flipda} propose a data augmentation method that combines prompting method with generating label-flipped data. To ease the manual effort of suitable prompt design, automatic prompt search has been extensively explored. \citet{shin2020autoprompt} explore gradient-guided search to generate both templates and label words. \citet{gao2021making} utilize sequence-to-sequence models to generate prompt candidates. These auto-generated \textit{hard} prompts cannot achieve competitive performance compared with manual prompts \citep{han2021ptr}. Thus, a series of research works on \textit{soft} prompts have been proposed, which directly use learnable continuous embeddings as prompt templates and work well on those large-scale PLMs \citep{li2021prefix,qin2021learning,lester2021power}.
Since the function of academic literature is relatively well-defined and does not require diversification, verbalizer mining methods \citep{hu2021knowledgeable,schick2020automatically,cui2022prototypical} are not considered in this paper.
We manual adopt the function labels and their related words as the verbalizer of prompts. This process does not require much human effort.
Moreover, we propose to combine the manual prompt with automatically learned continuous prompt which provides multi-perspective representations and takes full advantage of knowledge in the PLM.

\section{Task formulation and Datasets}
\subsection{Task formulation}
Our research objective is to use the text data of a scientific publication to recognize multi-granularity academic functions, e.g., the academic function of a citation, a abstract sentence, or a keyword. 
Formally, an academic function recognition training dataset can be denoted as $\mathcal{D} = \{\mathcal{X}, \mathcal{Y}\}$, where $\mathcal{X}$ is the instance set and $\mathcal{Y}$ is the academic function label set. Each instance $x\in \mathcal{X}$ consists of several tokens ${x}=(w_0, w_1,\dots,w_{|{x}|})$ along with a class label $y\in\mathcal{Y}$. The common approach is to train a model on the dataset $\mathcal{D}$. Whereas in real-world scenarios, annotated data for scientific NLP tasks are usually scarce. For instance, each class has only a few dozen or even about 10 labeled instances. Suppose there is a set of unlabeled instances $\mathcal{N}$, $\mathcal{|N|}$ is typically much larger than the number of training instances $\mathcal{|D|}$.
In this study, we explore an effective method to recognize multi-granularity academic function in low-resource settings. 

\subsection{Tasks and Corresponding Datasets}
To illustrate the effectiveness of MPT, we conduct extensive experiments on the following academic function recognition tasks of different granularities: (1) Citation Function Recognition; (2) Abstract Sentence Function Recognition; (3) Keyword Function Recognition. 

\textbf{Citation Function Recognition}. 
We carry out our experiment on a citation function recognition dataset, the \textbf{SciCite} \citep{cohan2019structural}, which is more than five times larger and covers multiple scientific domains compared with the ACL-ARC \citep{jurgens2016citation} dataset.
Specifically, there are 11,020 instances in the dataset extracted from 6,627 papers in the Computer Science domain and Medicine. Since they utilize a concise annotation scheme, only three types of function labels, i.e., ``\textit{Background}'', ``\textit{Method}'', and ``\textit{Result comparison}'', are assigned.

\textbf{Structure Function Recognition}. 
We evaluate our method on the medical scientific abstracts benchmark dataset \textbf{PubMed RCT 20k} \citep{dernoncourt2017pubmed} for abstract sentence function recognition, where each sentence of the abstract is annotated with one label associated with the rhetorical structural (``\textit{Background}'', ``\textit{Objective}'', ``\textit{Method}'', ``\textit{Result}'', and ``\textit{Conclusion}''). 

\textbf{Keyword Function Recognition}.
We conduct experiments of keyword function recognition task on \textbf{PMO-kw} dataset, which is a Chinese dataset proposed by \citet{zhang2021multiword} in computer science domain. PMO-kw contains 310,214 keywords extracted from 100,025 papers. Each keyword is annotated with one label associated with the domain-independent keyword semantic functions (``\textit{Problem}'', ``\textit{Method}'', and ``\textit{Others}'').

\begin{table}[!t]
  \centering
  \caption{Dataset statistics. CL denotes the computational linguistics domain. CS denotes the computer science domain. MED denotes the medicine domain. $|\mathcal{Y}$| means the label number of each dataset. \#Test means the instance number of test set.}
  \resizebox{0.45\textwidth}{!}{
    \begin{tabular}{lccccc}
    \toprule
    \textbf{Dataset} & \textbf{Language} & \textbf{Target} & \textbf{Domain} & \textbf{$|\mathcal{Y}|$} & \textbf{\#Test} \\
    \midrule
    \midrule
    SciCite & English & Citation Function & CS\&MED & 3     & 1,861 \\
    RCT-20k & English & Structure Function & MED   & 5     & 30,135 \\
    PMO-kw & Chinese & Keyword Function & CS    & 3     & 800 \\
    \bottomrule
    \end{tabular}%
    }
  \label{tab:dataset}%
\end{table}%

A summary of statistics of these tasks and datasets are shown in Table \ref{tab:dataset}.

\section{Methodology}
\subsection{Preliminaries}
\label{preliminary}
Before introducing our proposed Mix Prompt Tuning, we first give some essential preliminaries about prompt tuning for academic function recognition tasks.

\textbf{Fine-tuning}. Let $\mathcal{M}$ be a language model (PLM) pre-trained on large scale corpora. Previous works adopting the PLM to recognize the academic function mainly utilized the further pre-training and fine-tuning paradigms. \citet{gururangan2020don} find that the PLM could perform better by applying the self-supervised pre-training on the target domain corpus. For the fine-tuning paradigm, the instance ${x}$ is firstly converted to the sequence $($[\textit{CLS}]$, x_0, x_1,\dots,x_{|{x}|},$ [\textit{SEP}]$)$ by adding special tokens, [\textit{CLS}] and [\textit{SEP}], into it. Then $\mathcal{M}$ is used to encode the converted sequence into contextualized vectors $(\boldsymbol{h}_{[\text{\textit{CLS}}]}, \boldsymbol{h}_0, \boldsymbol{h}_1,\cdot,\boldsymbol{h}_{|{x}|}, \boldsymbol{h}_{[\text{\textit{SEP}}]})$. The mainstream methods all adopt the contextualized hidden state vector $\boldsymbol{h}_{[\text{\textit{CLS}}]}$ of $x_{[\text{\textit{CLS}}]}$ as the representation of the whole input sequence. Then a dense layer with learnable parameters $\boldsymbol{W}$ and $\boldsymbol{b}$ is utilized to estimate the probability distribution of the instance ${x}$ with a softmax function:
\begin{align}
    {\boldsymbol{p}}(\cdot|x)=\text{softmax}(\boldsymbol{W}\boldsymbol{h}_{[\text{\textit{CLS}}]}+\boldsymbol{b})
\end{align}
, where ${\boldsymbol{p}}\in \mathbb{R}^{|\mathcal{Y}|}$. 
The parameters of $\mathcal{M}$, $\boldsymbol{W}$, and $\boldsymbol{b}$ are tuned to minimize the loss $-\frac{1}{|\mathcal{X} |}\sum_{{x}\in\mathcal{X} }^{} log({\boldsymbol{p}}(y|x))$.

\textbf{Prompt-tuning}.
As mentioned above, fine-tuning requires learning extra classifiers on top of the PLM under different classification objectives, which needs more annotated data and makes the model hard to generalize well. 
Recently, a series of works \citep{radford2019language,petroni2019language,brown2020language} in general text classification tasks use prompt learning to bridge the gap between pre-training and downstream tasks, which make full use of PLMs by reformulating tasks as cloze-style objectives. 

Formally, a prompt $\mathcal{P}$ consists of a template $\mathcal{T}$, label words $\mathcal{V}$, and a verbalizer $\phi$. For each instance $x$, the template is leveraged to map $x$ to the prompt input $x_{prompt}=\mathcal{T}(x)$, which is named as template wrapping. The template $\mathcal{T}(\cdot)$ defines the location and number of added additional tokens. Taking the abstract sentence function recognition task as an example, we set the template $\mathcal{T}(\cdot)=$ ``$\cdot$ This sentence describes the [\textit{MASK}] of the paper.'', and map $x$ to $x_{prompt}=$ ``$x$ This sentence describes the [\textit{MASK}] of the paper.''. At least one [\textit{MASK}] is inserted into $x_{prompt}$ for the PLM $\mathcal{M}$ to fill the label words, while keeping the original tokens of $x$. 

After the template wrapping, we can obtain the hidden vector $\boldsymbol{h}_{[\text{\textit{MASK}}]}$ of the [\textit{MASK}] from $\mathcal{M}$ by encoding $x_{prompt}$.
Then we can utilize $\boldsymbol{h}_{[\text{\textit{MASK}}]}$ to produce a probability distribution $\boldsymbol{p}([\text{\textit{MASK}}]|x_{prompt})$ that reflects which tokens of $\mathcal{V}$ are suitable for replacing the [\textit{MASK}] token:
\begin{align}
    \boldsymbol{p}([\text{\textit{MASK}}]|x_{prompt})=\frac{exp(\boldsymbol{e}\cdot \boldsymbol{h}_{\text{[MASK]}})}{\sum_{v'\in\mathcal{V}}^{} exp(\boldsymbol{e'}\cdot \boldsymbol{h}_{\text{[MASK]}})} 
\end{align}
where $\boldsymbol{e}$ is the embedding of the token $v$ in $\mathcal{M}$.

\begin{figure*}[!t]
\centering
\includegraphics[width=6in]{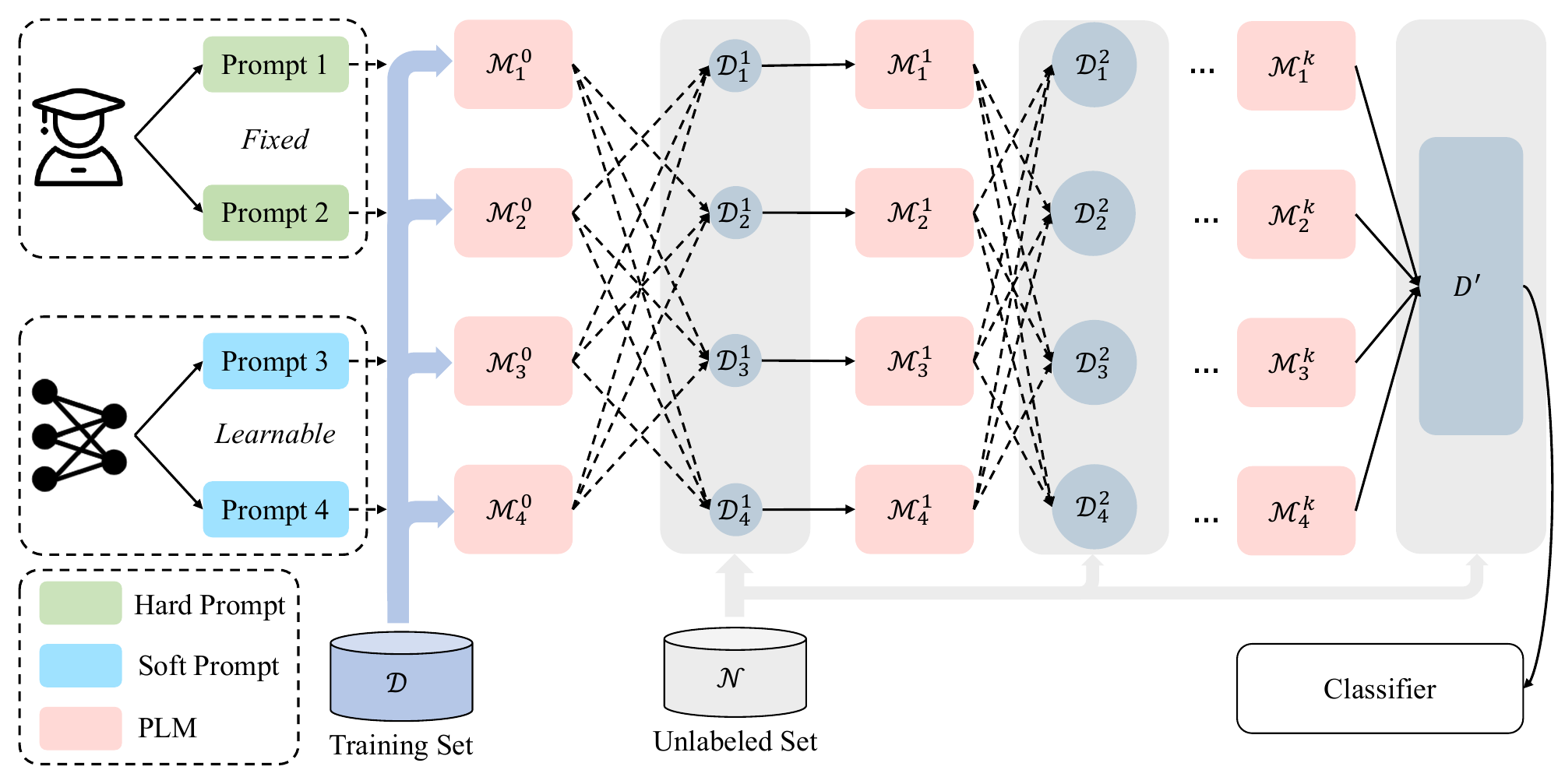}
\caption{The overview architecture of our proposed MPT.}
\label{fig_mpt}
\end{figure*}

In prompt learning, there is a \textit{verbalizer} as an injective mapping function $\phi:\mathcal{Y}\to \mathcal{V}$ that maps task labels to label words $\mathcal{V}$. For instance, we set $\phi(y=\text{``background''})\to\text{``literature''}$ and $\phi(y=\text{``method''})\to\text{``uses''}$. According to whether $\mathcal{M}$ predicts ``literature'' or ``uses'', we can know the function of the instance $x$ is ``background'' or ``method''. 
Note that for academic function recognition tasks, there may be a set of label token $\phi(y)=\mathcal{V}_y\subset\mathcal{V}$ that correspond to a particular label $y$. For instance, ``tool'', ``approach'', and ``method'' all indicate the ``method'' function in the citation function recognition task. 
Thus, the probability distribution of the [\textit{MASK}] over $\mathcal{Y}$ can obtained by $\boldsymbol{p}_{\mathcal{T}}(y|x) = \sum_{v\in\mathcal{V}_y}\boldsymbol{p}_{\mathcal{T}}([\text{\textit{MASK}}]=v|\mathcal{T}(x))$.
The parameters of $\mathcal{M}$ are tuned to minimize the loss $\mathcal{L}$:
\begin{align}
    \mathcal{L} = -\frac{1}{|\mathcal{X} |}\sum_{{x}\in\mathcal{X} }^{} log(\sum_{v\in\mathcal{V}_y}{\boldsymbol{p}}_{\mathcal{T}}(\text{[\textit{MASK}]}=v|\mathcal{T}(x)))
\end{align}

\subsection{Mix Prompt Tuning (MPT)}

So far, we have shown how to reformulate an academic function recognition task as a language modeling task using prompts. Here, we propose the \textit{Mix Prompt Tuning} (MPT) base on iPET \citep{schick2021exploiting} framework. The proposed semi-supervised method can not only combine the expert knowledge to design the appropriate manual prompt templates of academic function, but also adopt automatically learned \textit{soft} continuous prompt and manually designed \textit{hard} prompt templates. In this way, manually designed templates can improve the performance lower bound and maintain the performance stability. Moreover, automatically learned soft continuous prompt might elicit more knowledge using ``model's language'' to improve the performance upper bound. As for the verbalizer $\phi$, since each class in an academic function recognition task is clearly defined, in this paper, we utilize the same one verbalizer for all of the templates in one task.

Suppose we have a set of unlabeled instances $\mathcal{N}$, which is typically larger than the instance number of training set $\mathcal{D}$.
First, we define a set $\Gamma$ of templates, which contains learnable \textit{soft} continuous prompt templates, i.e., Prompt 3 and Prompt 4 in Figure \ref{fig_mpt}, and manual designed fixed \textit{hard} prompt templates, i.e., Prompt 1 and Prompt 2 in Figure \ref{fig_mpt}. Then we fine-tune the PLM $\mathcal{M}$ based on one prompt template $\mathcal{T}\in\Gamma$ and the fixed verbalizer $\phi$ in the prompt tuning manner. With the original training instances $\mathcal{X}$ and the unlabeled instances $\mathcal{N}$, multiple PLMs tuned on prompt set can produce \textbf{pseudo labeled training set} by:
\begin{align}
    \boldsymbol{p}_{\mathcal{M}}(y|x) = \frac{1}{Z}\sum_{\mathcal{T}\in{\Gamma}}^{} \omega(\mathcal{T})\cdot \boldsymbol{p}_{\mathcal{T}}(y|x)
    \label{formula:pseudo}
\end{align}
where $Z=\sum_{\mathcal{T}\in{\Gamma}}^{} \omega(\mathcal{T})$ and $\omega(\mathcal{T})$ denotes the weight of template $\mathcal{T}$.
The pseudo labeled training set is finally utilized to train a classifier. 

Since there is high probability that PLMs tuned on some prompt templates perform worse than PLMs tuned on other prompt templates, to force them learn from each other, we further train several generations of models using unlabeled dataset and knowledge distillation strategy. Specifically, we formally denote the first generation PLMs tuned on $\mathcal{D}$ as $\mathcal{M}^{0}=\{\mathcal{M}_{1}^{0},\dots, \mathcal{M}_{i}^{0},\dots,\mathcal{M}_{|\Gamma|}^{0}\}$, where $\mathcal{M}_{i}^{0}$ is tuned with template $\mathcal{T}_i$ and verbalizer $\phi$. After $k$ generations training, we can get models $\mathcal{M}^{1},\dots,\mathcal{M}^{j},\dots,\mathcal{M}^{k}$, where $\mathcal{M}^{j}=\{\mathcal{M}_{1}^{j},\dots, \mathcal{M}_{i}^{j},\dots,\mathcal{M}_{|\Gamma|}^{j}\}$ and $\mathcal{M}_{i}^{j}$ is trained with $\mathcal{T}_{i}$ on training dataset $\mathcal{D}_{i}^{j}$. Note that we need to keep the proportion of classes in the dataset constant to avoid learning a biased distribution with iteration. Thus, we expand the size of the labeled dataset by a constant factor $d\in\mathbb{N}$, i.e., $c_j(y)=d\cdot c_{j-1}(y)$, where $c_j(y)$ denotes the count of instances with label $y$ in generation $j$ using template set $\Gamma$.

To obtain the training dataset $\mathcal{D}_{i}^{j}$, we randomly sample $\lambda \cdot (|\Gamma| -1)$ models $\mathcal{S}$ from previous generation $j-1$ tuned PLMs $\mathcal{M}^{j-1}$ except $\mathcal{M}_{i}^{j-1}$, where $\lambda\in(0,1]$. With the subset models $\mathcal{S}$, the unlabeled data can be annotated by formula \ref{formula:pseudo} to construct a labeled dataset $\mathcal{D}_{\mathcal{S}}$:
\begin{align}
    \mathcal{D}_{\mathcal{S}}=\{(x,\arg\max_{y\in\mathcal{Y}} \boldsymbol{p}_{S}(y|x) | x\in\mathcal{N} ) \}
\end{align}
Inspired by the findings \citep{guo2017calibration} that instances predicted with high confidence are typically more likely to be classified correctly. As a result, we sample the instances with high probability score of the labels to avoid training next generation model on mislabeled data and improve the performance lower bound. Formally, for class $y\in\mathcal{Y}$, $\mathcal{D}_\mathcal{S}(y)\subset\mathcal{D}_\mathcal{S}$ is constructed by choosing the top $c_{j}(y)-c_{0}(y)$ scores of label $y$ instances from $\mathcal{D}_{\mathcal{S}}$. , the training dataset $\mathcal{D}_{i}^{j}$ is composed of the initial training set $\mathcal{D}$ and \textbf{even sampled} pseudo labeled training set $\bigcup_{y\in\mathcal{Y} }\mathcal{D}_{\mathcal{S}}(y)$, i.e., $\mathcal{D}_{i}^{j}=\mathcal{D}\cup\bigcup_{y\in\mathcal{Y} }\mathcal{D}_{\mathcal{S}}(y)$. The final enlarged dataset $\mathcal{D}'$, annotated by the last generation models $\mathcal{M}^k$, is adopted to train a classifier.

As $\mathcal{D}_{\mathcal{S}}$ may not contain enough instances for a class $y$ under the extremely unbalanced label distribution, we obtain all $\mathcal{D}_{\mathcal{S}}(y)$ by choosing the highest $\boldsymbol{p}_{\mathcal{S}}(y|x)$ instances. 

\subsection{Designing Template and Verbalizer}
\begin{table}[!t]
  \centering
  \caption{Prompt templates and verbalizers. Note that the templates and verbalizer for the \textit{Keyword Function} recognition task are translated from Chinese. }
  \resizebox{0.48\textwidth}{!}{
    \begin{tabular}{lll}
    \toprule
    \textbf{Task} & \textbf{Prompt Template} & \textbf{Verbalizer} \\
    \midrule
    \midrule
    \multirow{2}[2]{*}{All} & $\mathcal{T}_1(x)$ = $x$. <Soft> <Soft> [MASK] & \multicolumn{1}{l}{\multirow{2}[2]{*}{-}} \\
          & $\mathcal{T}_2(x)$ = $x$. <Soft> <Soft> <Soft> [MASK] & \\
    \midrule
    \multirow{4}[2]{*}{\makecell{Citation\\Function}} & $\mathcal{T}_1(x)$ = <Task Description> $x$. Citation Function: [MASK] & \multirow{4}[2]{*}{\makecell[l]{Background: [background, literature]\\Method: [method, approach]\\Result: [result]}} \\
          & \makecell[tl]{$\mathcal{T}_2(x)$ = <Task Description> $x$. The function of this citation \\\qquad\qquad is [MASK]} &  \\
          & $\mathcal{T}_3(x)$ = $x$. The function of this citation is [MASK] &  \\
          & $\mathcal{T}_4(x)$ = $x$. Citation Function: [MASK] &  \\
    \midrule
    \multirow{4}[2]{*}{\makecell{Structure\\Function}} & $\mathcal{T}_1(x)$ = <Task Description> $x$. Structure Function: [MASK] & \multirow{4}[2]{*}{\makecell[l]{Background: [background]\\Objective: [objective]\\Methods: [methods]\\Results: [results]\\Conclusions: [conclusions]}} \\
          & \makecell[l]{$\mathcal{T}_2(x)$ = <Task Description> $x$. The structure function of \\\qquad\qquad this sentence is [MASK]} &  \\
          & $\mathcal{T}_3(x)$ = $x$. The Structure function of this sentence is [MASK] &  \\
          & $\mathcal{T}_4(x)$ = $x$. Structure Function: [MASK] &  \\
    \midrule
    \multirow{4}[2]{*}{\makecell{Keyword\\Function}} & \makecell[tl]{$\mathcal{T}_1(x)$ = <MASK> is the function of $x$ in <Abstract>.\\\qquad\qquad <Title>. <Task Description> } & \multicolumn{1}{l}{\multirow{4}[2]{*}{\makecell[l]{Method: [method, algorithm, technology]\\Problem: [problem, target, orientation]\\Others: [data, metric, tool]}}} \\
          & \makecell[tl]{$\mathcal{T}_2(x)$ = Keyword function: <MASK>. $x$ in <Abstract>.\\\qquad\qquad<Title>. <Task Description>} &  \\
          & $\mathcal{T}_3(x)$ = <MASK> is the function of $x$ in <Abstract>. &  \\
          & $\mathcal{T}_4(x)$ = Keyword function: <MASK>. $x$ in <Abstract>. &  \\
    \bottomrule
    \end{tabular}%
    }
  \label{tab:template}%
\end{table}%

We now describe the templates and verbalizers adopted for each task. For all tasks, we add two or three soft learnable continuous tokens between the instance content and the [\textit{MASK}] as \textit{soft} prompt templates. Moreover, we manually design four \textit{hard} templates for each task by combing the domain knowledge. Two of these manually designed \textit{hard} templates are added with the task description. Specifically, for citation function recognition, we adopt ``\textit{Citation function identifies the meaning or purpose behind a particular citation.}'' \citep{pride2020authoritative} as the description and add it at the beginning of $\mathcal{T}_1$ and $\mathcal{T}_2$. For structure function recognition, we adopt ``\textit{An abstract is divided into semantic headings such as background, objective, method, result, and conclusion.}'' \citep{dernoncourt2016neural}. For keyword function recognition, we adopt ``\textit{The functions carried by the keywords in the literature are problem, method and others (in Chinese).}'' \citep{wei2020recognition}. Since each class in an academic function recognition is clearly defined and does not require diversification, we construct the verbalizer of labels with respect to a class according to the class definition or description that reflect the expert knowledge. This process does not require much human effort. More details of prompt templates and verbalizers for all tasks are shown in Table \ref{tab:template}.

\section{Experiments}
Scientific NLP tasks are considered difficult not only because of the task itself, but also because of data scarcity. For many scientific NLP tasks, since task-specific annotations are difficult to obtain, we only have access to a limited amount of annotated data. Therefore, following previous low-resource work in other domains \citep{xie2020unsupervised,chen2020mixtext,schick2021exploiting}, we used only 4 to 128 samples per class to construct low-resource scenarios to further increase the difficulty and align with real-world practical situations.

\subsection{Baselines}
To test the effectiveness of our method, we compare it with several types of recent models:

\textit{Supervised baselines.} We compare several fine-tuning and prompt-tuning baselines in supervised manner, which have been proved the effectiveness on general text classification tasks in low-resource setting. 

\textbf{Fine-tuning.} As PLMs have achieved promising results on various NLP tasks, a lot of efforts have been devoted to fine-tuning PLMs for text classification as well. In this paper, we select (1) \textbf{BERT} \citep{devlin2019bert}, (2) \textbf{RoBERTa} \citep{liu2019roberta}, and (3) \textbf{SciBERT} \citep{beltagy2019scibert} as the representative fine-tuning baselines. For fair comparison, we choose the base version of these models, e.g., BERT-base-uncased. 

\textbf{Prompt-tuning.} We apply the regular prompt-tuning paradigm (described in subsection \ref{preliminary}) with the \textit{hard} and \textit{soft} prompt templates (shown in Table \ref{tab:template}) to form the (4) \textbf{PT-hard} and (5) \textbf{PT-soft} models, respectively. Since there are multiple templates, we tune the PLM with different prompt templates and report the best performance.

\textit{Semi-supervised and data augmentation baselines.} Semi-supervised learning has been widely used in different NLP tasks combining with massive amount of unlabeled data to improve the model performance, as unlabeled data is often plentiful compared to labeled data. In this paper, we adopt five classic or previous SOTA methods for semi-supervised learning in NLP that rely on data augmentation: (6) Unsupervised Data Augmentation (\textbf{UDA}) \citep{xie2020unsupervised} connects data augmentation with semi-supervised learning and outperforms previous SOTA. It employs back translation and TF-IDF to generate diverse and realistic noise and enforces the model to be consistent with respect to these noise. (7) \textbf{TMix} \citep{chen2020mixtext} proposes to create virtual training samples by apply linear interpolations within hidden space, produced by PLMs, as a data augmentation method for text. (8) \textbf{MixText} \citep{chen2020mixtext} leverages TMix both on labeled and unlabeled data for semi-supervised learning. The authors propose the label guessing method to generate labels for the unlabeled data in the training process. (9) \textbf{PET} \citep{schick2021exploiting} is a semi-supervised training procedure that reformulates input examples as cloze-style phrases by prompt templates and verbalizers to help leverage the knowledge contained in PLMs. Along with PLMs, these well-designed templates are used to assign soft labels to unlabeled examples. A standard classifier is trained based on the original labeled data and soft labeled data. (10) \textbf{iPET} \citep{schick2021exploiting} is the iterative variant of PET. It trains several generations of models using PET manner on datasets of increasing size so that classifiers can learn from each other.
We adjust the open-sourced implementations of UDA \footnote{https://github.com/SanghunYun/UDA\_pytorch}, TMix, MixText\footnote{https://github.com/GT-SALT/MixText}, PET, and iPET\footnote{https://github.com/timoschick/pet} to conduct academic function recognition tasks on corresponding datasets.

\subsection{Experimental Settings} 
Since label distributions of academic function recognition datasets are imbalanced, we conduct comparative experiments on balanced and imbalanced label distributions under the low-resource setting. Previous baselines, such as MixText and PET, only conduct the experiments on balanced sampled training set. Moreover, taking the unlabeled dataset of MixText as an example, it is unrealistic and impractical that the unlabeled dataset is carefully selected and balanced sampled. Thus, to conduct the experiments on balanced label distribution, we randomly sample $K=\{4,8,16,32,64,128\}$ instances in each class from the training set and test the model on the entire test set. Whereas, for the experiments on imbalanced label distribution, i.e., original label distribution, we randomly sample the same number of training instances ($K*|\mathcal{Y}|$) with the experiments on balanced label distribution by keeping the original label distribution.
For all datasets, we use the macro F1 score and the Accuracy. 

All our models and baselines are implemented with PyTorch framework \citep{paszke2019pytorch} and Huggingface transformers \citep{wolf2020transformers}. All of the models related to prompt learning are also implemented with the Open-Prompt toolkit \citep{ding2021openprompt}. We fine-tune the PLMs with the AdamW optimizer \citep{loshchilov2018decoupled}. Previous study \citep{gao2021making} find that, with a large validation set, a model could learn more knowledge and hyperparameters could also be optimized. Different from the experimental settings of \citet{xie2020unsupervised} and \citet{chen2020mixtext} that adopt large validation sets to optimize the hyperparameters, to keep the initial goal of learning from limited data, we assume that there is no access to a large validation set and the size of validation set is the same as training set. 
For citation function recognition task and structure function recognition task, we use SciBERT \citep{beltagy2019scibert} as our PLM backbone. \textbf{Whereas for keyword function recognition task, since there is only Chinese dataset, we use bert-base-multilingual-cased\footnote{https://huggingface.co/bert-base-multilingual-cased} as our PLM backbone.}
We use a learning rate of $1\cdot10^{-5}$, a batch size of 16, a maximum sequence length of 128 for citation function recognition and structure function recognition, and a maximum sequence length of 256 for keyword function recognition. Same with the PET framework, we set $\lambda=0.25$ and $d=5$. 
For the comparative experiment of semi-supervised methods, the number of unlabeled instances is selected from $\{600, 800, 1000\}$.
We tune the entire model for 6 epochs under 3 different random seeds and report the best test performance.

\subsection{Main Results}

In this subsection, we introduce the specific comparative results and provide possible insights of our proposed MPT under two settings, i.e., balanced sample $K$ instances ($K$-shot) in each class and randomly sample $K*|\mathcal{Y}|$ instances from the original training set. 
\label{results}

\subsubsection{Balanced sample few-shot}

\textbf{Few-shot supervised methods.} The experimental results of comparisons with supervised base model under balanced label distribution are shown in Table \ref{tab:bal_sup}. We can observe that:

\begin{table}[!htbp]
  \centering
  \caption{Few-shot experimental results of performance (\%) comparison with supervised base model on RCT-20k, SciCite, and PMO-kw test sets. Note that the training data are balanced constructed by sampling $K$ instances in each class from the original training set, $K=\{4,8,16,32,64,128\}$. \textbf{\textit{Bold}} shows the best performance corresponding to $K$.}
  \resizebox{0.49\textwidth}{!}{
    \begin{tabular}{clcccccc}
    \toprule
    \multirow{2}[4]{*}{\# Balanced $K$} & \multicolumn{1}{c}{\multirow{2}[4]{*}{Method}} & \multicolumn{2}{c}{RCT-20k} & \multicolumn{2}{c}{SciCite} & \multicolumn{2}{c}{PMO-kw} \\
\cmidrule{3-8}          &       & Accuracy & Macro F1 & Accuracy & Macro F1 & Accuracy & Macro F1 \\
    \midrule
    \midrule
    \multirow{6}[2]{*}{4} & BERT  & 39.10 & 31.62 & 23.05 & 18.88 & 25.38 & 9.97 \\
          & RoBERTa & 31.34 & 23.68 & 43.20 & 36.04 & 38.13 & 28.74 \\
          & SciBERT & 66.12 & 58.96 & 51.37 & 51.03 & 35.13 & 33.81 \\
          & PT-soft & 38.95 & 31.95 & 33.96 & 27.16 & 33.50 & 28.68 \\
          & PT-hard & 35.03 & 30.97 & 45.73 & 29.68 & 30.00 & 19.10 \\
          & MPT   & \textbf{78.82} & \textbf{73.34} & \textbf{82.91} & \textbf{78.69} & \textbf{46.37} & \textbf{46.32} \\
    \midrule
    \multirow{6}[2]{*}{8} & BERT  & 57.60 & 51.63 & 46.96 & 41.41 & 44.00 & 26.15 \\
          & RoBERTa & 32.83 & 24.21 & 41.05 & 35.17 & 39.00 & 33.87 \\
          & SciBERT & 69.76 & 63.12 & 63.30 & 62.34 & 38.63 & 38.18 \\
          & PT-soft & 57.36 & 49.39 & 36.33 & 28.39 & 50.13 & 50.11 \\
          & PT-hard & 54.32 & 48.64 & 60.25 & 50.02 & 39.88 & 30.60 \\
          & MPT   & \textbf{81.19} & \textbf{75.76} & \textbf{83.93} & \textbf{82.40} & \textbf{55.50} & \textbf{55.11} \\
    \midrule
    \multirow{6}[1]{*}{16} & BERT  & 61.39 & 56.16 & 64.32 & 56.82 & 30.00 & 28.14 \\
          & RoBERTa & 42.96 & 39.95 & 41.00 & 39.50 & 35.75 & 33.96 \\
          & SciBERT & 66.30 & 60.40 & 75.50 & 74.02 & 34.50 & 31.49 \\
          & PT-soft & 70.31 & 63.03 & 82.43 & 80.50 & 64.75 & 64.73 \\
          & PT-hard & 69.79 & 63.27 & 70.67 & 62.81 & 48.63 & 45.97 \\
          & MPT   & \textbf{82.17} & \textbf{76.48} & \textbf{86.57} & \textbf{84.32} & \textbf{71.25} & \textbf{71.08} \\
    \midrule
    \multirow{6}[1]{*}{32} & BERT  & 68.67 & 62.59 & 77.43 & 74.81 & 47.00 & 46.96 \\
          & RoBERTa & 61.08 & 52.58 & 76.25 & 73.53 & 39.50 & 37.00 \\
          & SciBERT & 74.81 & 68.06 & 81.03 & 79.65 & 51.63 & 51.04 \\
          & PT-soft & 75.22 & 69.24 & \textbf{85.65} & \textbf{84.14} & 73.75 & 72.70 \\
          & PT-hard & 72.43 & 66.57 & 76.21 & 69.12 & 68.63 & 68.52 \\
          & MPT   & \textbf{81.94} & \textbf{76.23} & 83.02 & 81.27 & \textbf{78.25} & \textbf{78.35} \\
    \midrule
    \multirow{6}[2]{*}{64} & BERT  & 73.59 & 67.94 & 79.21 & 76.77 & 72.50 & 71.70 \\
          & RoBERTa & 72.58 & 66.08 & 78.02 & 76.06 & 47.38 & 45.62 \\
          & SciBERT & 77.57 & 70.03 & 84.42 & 82.87 & 83.63 & 83.31 \\
          & PT-soft & 75.97 & 70.58 & \textbf{86.51} & \textbf{84.86} & 83.63 & 83.69 \\
          & PT-hard & 79.76 & 73.54 & 78.95 & 71.82 & 82.50 & 82.44 \\
          & MPT   & \textbf{83.34} & \textbf{77.45} & 84.52 & 82.55 & \textbf{85.37} & \textbf{85.34} \\
    \midrule
    \multirow{6}[2]{*}{128} & BERT  & 76.91 & 70.73 & 83.24 & 81.16 & 82.00 & 81.46 \\
          & RoBERTa & 73.59 & 67.54 & 82.27 & 80.16 & 82.63 & 82.54 \\
          & SciBERT & 80.84 & 75.00 & 84.58 & 82.98 & 86.63 & 86.44 \\
          & PT-soft & 79.72 & 73.34 & \textbf{86.57} & \textbf{85.20} & 90.13 & 89.82 \\
          & PT-hard & 79.63 & 73.81 & 81.00 & 75.20 & 87.88 & 87.55 \\
          & MPT   & \textbf{82.44} & \textbf{76.49} & 84.58 & 82.99 & \textbf{90.25} & \textbf{90.01} \\
    \bottomrule
    \end{tabular}%
    }
  \label{tab:bal_sup}%
\end{table}%

(1) Domain pre-trained model, i.e., SciBERT, achieves better overall performance than other models pre-trained on non-scientific-corpus, even though these non-scientific-corpus are much larger than the pre-trained corpus of SciBERT (approximately 13GB of SciBERT vs. 160GB of RoBERTa). It indicates that the performance gains are from scientific-domain data, which may also contribute to the performance improvement of prompt-learning based method. We will further discuss it in the following Section \ref{sec:analyze}.

\begin{table}[!htbp]
  \centering
  \caption{Few-shot experimental results of performance (\%) comparison with semi-supervised base models on RCT-20k, SciCite, and PMO-kw test sets. Note that the training data are balanced constructed by sampling $K$ instances in each class from the original training set, where $K=\{4,8,16,32,64,128\}$. \textbf{\textit{Bold}} shows the best performance corresponding to $K$.}
  \resizebox{0.48\textwidth}{!}{
    \begin{tabular}{clcccccc}
    \toprule
    \multirow{2}[4]{*}{\# Balanced $K$} & \multicolumn{1}{c}{\multirow{2}[4]{*}{Method}} & \multicolumn{2}{c}{RCT-20k} & \multicolumn{2}{c}{SciCite} & \multicolumn{2}{c}{PMO-kw} \\
\cmidrule{3-8}          &       & Accuracy & Macro F1 & Accuracy & Macro F1 & Accuracy & Macro F1 \\
    \midrule
    \midrule
    \multirow{6}[2]{*}{4} & TMix  & 37.35 & 34.32 & 36.38 & 34.89 & 36.50 & 35.69 \\
          & UDA   & 57.89 & 47.88 & 54.55 & 52.38 & 27.27 & 14.29 \\
          & MixText & 36.86 & 35.72 & 39.66 & 36.77 & 36.88 & 35.85 \\
          & PET   & 74.18 & 68.51 & 42.34 & 42.02 & 38.63 & 38.48 \\
          & iPET  & \textbf{78.87} & 72.73 & 62.76 & 62.00 & \textbf{46.38} & \textbf{46.45} \\
          & MPT   & 78.82 & \textbf{73.34} & \textbf{82.91} & \textbf{78.69} & 46.37 & 46.32 \\
    \midrule
    \multirow{6}[2]{*}{8} & TMix  & 44.74 & 39.28 & 44.49 & 42.50 & 38.38 & 34.36 \\
          & UDA   & 69.23 & 69.88 & 78.26 & 78.24 & 39.13 & 31.28 \\
          & MixText & 51.71 & 47.70 & 54.43 & 45.06 & 39.50 & 35.71 \\
          & PET   & 79.56 & 73.79 & 53.73 & 54.68 & 47.63 & 47.79 \\
          & iPET  & 79.80 & 74.00 & 68.51 & 67.01 & 46.75 & 46.80 \\
          & MPT   & \textbf{81.19} & \textbf{75.76} & \textbf{83.93} & \textbf{82.40} & \textbf{55.50} & \textbf{55.11} \\
    \midrule
    \multirow{6}[2]{*}{16} & TMix  & 55.47 & 49.94 & 59.81 & 52.26 & 41.38 & 33.42 \\
          & UDA   & 77.22 & 77.49 & 74.47 & 72.78 & 29.79 & 27.78 \\
          & MixText & 55.80 & 50.05 & 69.59 & 64.90 & 43.13 & 34.83 \\
          & PET   & 80.32 & 74.63 & 72.17 & 70.81 & 48.50 & 48.09 \\
          & iPET  & 77.58 & 71.86 & 76.09 & 74.50 & 46.25 & 46.17 \\
          & MPT   & \textbf{82.17} & \textbf{76.48} & \textbf{86.57} & \textbf{84.32} & \textbf{71.25} & \textbf{71.08} \\
    \midrule
    \multirow{6}[2]{*}{32} & TMix  & 58.01 & 52.89 & 71.57 & 68.05 & 40.50 & 37.78 \\
          & UDA   & 77.99 & 77.67 & 74.74 & 74.76 & 32.63 & 32.22 \\
          & MixText & 64.83 & 60.91 & 77.86 & 75.68 & 38.13 & 36.44 \\
          & PET   & 80.15 & 74.00 & 76.30 & 74.94 & 57.00 & 56.52 \\
          & iPET  & 78.22 & 71.92 & 78.24 & 76.71 & 53.38 & 52.94 \\
          & MPT   & \textbf{81.94} & \textbf{76.23} & \textbf{83.02} & \textbf{81.27} & \textbf{78.25} & \textbf{78.35} \\
    \midrule
    \multirow{6}[2]{*}{64} & TMix  & 76.72 & 69.67 & 74.85 & 72.25 & 40.38 & 36.72 \\
          & UDA   & 77.43 & 77.45 & 81.15 & 80.92 & 37.70 & 37.70 \\
          & MixText & 77.83 & 71.59 & 80.44 & 77.77 & 42.00 & 39.47 \\
          & PET   & 81.75 & 75.55 & 79.80 & 78.16 & 66.13 & 65.91 \\
          & iPET  & 80.92 & 74.37 & 80.06 & 78.40 & 63.00 & 62.65 \\
          & MPT   & \textbf{83.34} & \textbf{77.45} & \textbf{84.52} & \textbf{82.55} & \textbf{85.37} & \textbf{85.34} \\
    \midrule
    \multirow{6}[2]{*}{128} & TMix  & 78.00 & 71.69 & 80.71 & 78.90 & 52.38 & 51.97 \\
          & UDA   & 75.59 & 75.53 & 79.63 & 79.22 & 41.78 & 41.45 \\
          & MixText & 77.28 & 71.89 & 83.13 & 81.03 & 50.25 & 50.34 \\
          & PET   & 81.99 & 75.82 & 81.52 & 79.64 & 65.63 & 65.52 \\
          & iPET  & 81.72 & 75.53 & 82.70 & 80.94 & 64.63 & 64.57 \\
          & MPT   & \textbf{82.44} & \textbf{76.49} & \textbf{84.58} & \textbf{82.99} & \textbf{90.25} & \textbf{90.01} \\
    \bottomrule
    \end{tabular}%
    }
  \label{tab:bal_semi}%
\end{table}%

(2) As the labeled data increases, prompt-tuning based methods gradually perform better than fine-tuning based methods. Moreover, prompt-tuning based methods also show greater performance gains than fine-tuning based methods. For instance, \textbf{PT-soft} performs better than other PLM based fine-tuning models when $K\ge16$ on RCT-20k, SciCite, and PMO-kw. This indicates that prompt-tuning based method can effectively stimulate the ability of the PLM and make full use of the scientific knowledge containing in pre-train corpus.

(3) The proposed method performs much better than the fully supervised baselines. Besides the contribution of SciBERT or multilingual backbone, MPT makes full use of the additional unlabeled data with the help of interactive learning from multiple PLMs and pre-train scientific knowledge stimulated by the prompt templates.

(4) With the sampling number $K$ increases, all of the recognition performance gradually improves. Moreover, MPT performs significantly higher than the baseline models in most of settings and datasets. For instance, for the PMO-kw dataset, MPT achieves 46.32\% Macro F1 score when $K=4$, which is significantly better than most of baselines. When $K$ increases to 32, PT-soft achieves 72.7\%. The performance is nearly doubled, but it still cannot surpass MPT, which achieves a significantly better performance of 78.35\%.

\textbf{Few-shot semi-supervised methods.} 
We also demonstrate the comparative experimental results of with semi-supervised baselines under balanced label distribution (few-shot setting) are shown in Table \ref{tab:bal_semi}. We can observe that: 

(1) Overall, semi-supervised methods perform better than fine-tuning based methods. It is because semi-supervised methods utilize back-translation or prompt-tuning to introduce more knowledge into the training phase. Back-translation is a data augmentation method that generates different instances integrated with language diversity and model knowledge. Back-translation based methods shows the superiority under extreme few training instances. However, with the sampling number $K$ increases, fine-tuning based methods performs better than back-translation based methods gradually. By analyzing the translation instances, it is found that there are many samples with poor translation quality, which introduces noise and affect the model performance as the number increases.

(2) Prompt-based semi-supervised methods performs better than other semi-supervised methods. In general, PET, iPET, and our proposed MPT performs better than other back-translation based semi-supervised methods in all $K$-shot settings. The performance of back-translation based methods are highly relevant to the translation quality. Compared with prompt-tuning based methods, back-translation based methods cannot stimulate the knowledge of language models directly and effectively.

(3) MPT shows the superiority over other baseline methods under low-resource settings on three datasets. With the same contribution of SciBERT or multilingual BERT, it demonstrates the effectiveness of MPT, which make full use of soft and hard prompt templates to obtain pseudo labels from unlabeled data and force multiple PLMs learn from each other interactively.

\subsubsection{Randomly sample $K*|\mathcal{Y}|$ instances}
Since previous comparative experiments are under few-shot settings that balanced sample $K$ instances from the original training set, we also conduct extensive experiments to evaluate the performance of different methods with the original class distribution. 
Overall, there is some degree of improvement or degradation in the performance of different models. For instance, for the SciCite dataset, MPT trained with few-shot settings performs better than the MPT trained with random samples when data resources are extremely scarce. However, with the growth of data volume, MPT trained with random samples shows a better performance upper boundary, compared to MPT trained with few-shot settings. It is because that models have more opportunity to see the samples of scarce categories in the balanced label distribution of few-shot setting. In the case where the amount of data is not extremely scarce, the consistency between the training distribution and the original distribution will largely affect the performance upper bound.
Despite this, Table \ref{tab:unbal_sup} and Table \ref{tab:unbal_semi} show that MPT substantially outperforms most of the baselines across all tasks for different training sizes.

\begin{table}[!htbp]
  \centering
  \caption{Low-resource experimental results of performance (\%) comparison with fully supervised base models on RCT-20k, SciCite, and PMO-kw test sets. Note that the training data are constructed by randomly sampling $K*|\mathcal{Y}|$ instances from the original training set, where $K=\{4,8,16,32,64,128\}$. \textbf{\textit{Bold}} shows the best performance corresponding to $K$.}
  \resizebox{0.48\textwidth}{!}{
    \begin{tabular}{clcccccc}
    \toprule
    \multirow{2}[4]{*}{\# Examples$=K*|\mathcal{Y}|$} & \multicolumn{1}{c}{\multirow{2}[4]{*}{Method}} & \multicolumn{2}{c}{RCT-20k} & \multicolumn{2}{c}{SciCite} & \multicolumn{2}{c}{PMO-kw} \\
\cmidrule{3-8}          &       & Accuracy & Macro F1 & Accuracy & Macro F1 & Accuracy & Macro F1 \\
    \midrule
    \midrule
    \multirow{6}[2]{*}{|D|=4*$|\mathcal{Y}|$} & BERT  & 45.01 & 29.63 & 54.00 & 37.58 & 30.63 & 17.54 \\
          & RoBERTa & 32.84 & 9.89  & 13.92 & 8.15  & 40.00 & 27.77 \\
          & SciBERT & 58.37 & 39.04 & 67.28 & \textbf{50.87} & 41.25 & 28.80 \\
          & PT-soft & 48.01 & 36.30 & 60.18 & 39.87 & 27.25 & 24.28 \\
          & PT-hard & 34.94 & 17.03 & 52.82 & 25.89 & 42.38 & 31.39 \\
          & MPT   & \textbf{69.97} & \textbf{50.78} & \textbf{64.59} & 42.74 & \textbf{45.00} & \textbf{42.11} \\
    \midrule
    \multirow{6}[2]{*}{|D|=8*$|\mathcal{Y}|$} & BERT  & 57.82 & 38.65 & 42.83 & 33.61 & 33.88 & 29.28 \\
          & RoBERTa & 33.02 & 17.80 & 36.49 & 28.90 & 35.88 & 29.66 \\
          & SciBERT & 68.07 & 50.81 & 64.97 & 51.62 & 34.88 & 30.88 \\
          & PT-soft & 53.43 & 40.56 & 59.81 & 40.69 & 46.25 & 38.39 \\
          & PT-hard & 73.37 & \textbf{66.29} & 53.04 & 25.79 & 45.63 & 33.38 \\
          & MPT   & \textbf{76.41} & 64.44 & \textbf{77.32} & \textbf{55.77} & \textbf{58.87} & \textbf{57.45} \\
    \midrule
    \multirow{6}[2]{*}{|D|=16*$|\mathcal{Y}|$} & BERT  & 64.56 & 56.04 & 53.57 & 36.19 & 27.50 & 17.47 \\
          & RoBERTa & 51.95 & 36.01 & 48.25 & 37.83 & 39.38 & 34.62 \\
          & SciBERT & 72.96 & 60.79 & 78.13 & 73.61 & 32.38 & 30.83 \\
          & PT-soft & 67.43 & 60.29 & 78.67 & 75.34 & 65.75 & 64.31 \\
          & PT-hard & 71.91 & 64.21 & 83.34 & \textbf{79.13} & 50.13 & 45.31 \\
          & MPT   & \textbf{82.43} & \textbf{75.32} & \textbf{84.09} & 78.91 & \textbf{68.37} & \textbf{66.50} \\
    \midrule
    \multirow{6}[2]{*}{|D|=32*$|\mathcal{Y}|$} & BERT  & 70.46 & 59.87 & 70.50 & 56.45 & 41.00 & 34.05 \\
          & RoBERTa & 70.51 & 61.85 & 50.94 & 30.90 & 38.38 & 32.35 \\
          & SciBERT & 80.95 & 72.86 & 80.12 & 75.03 & 59.63 & 57.34 \\
          & PT-soft & 75.07 & 68.51 & 82.27 & 78.42 & \textbf{75.88} & \textbf{73.76} \\
          & PT-hard & 76.30 & 70.23 & 85.22 & 83.05 & 74.38 & 71.93 \\
          & MPT   & \textbf{82.89} & \textbf{74.73} & \textbf{87.91} & \textbf{86.55} & 73.00 & 72.78 \\
    \midrule
    \multirow{6}[2]{*}{|D|=64*$|\mathcal{Y}|$} & BERT  & 75.50 & 66.22 & 80.06 & 76.51 & 58.00 & 56.17 \\
          & RoBERTa & 76.32 & 69.63 & 83.72 & 81.85 & 39.13 & 38.76 \\
          & SciBERT & 81.27 & 73.81 & 86.19 & 84.75 & 66.13 & 66.18 \\
          & PT-soft & 79.69 & 73.04 & 84.26 & 82.55 & 81.38 & 80.96 \\
          & PT-hard & 77.79 & 70.56 & 85.65 & 84.24 & 84.50 & 84.39 \\
          & MPT   & \textbf{83.71} & \textbf{75.43} & \textbf{88.18} & \textbf{86.90} & \textbf{85.62} & \textbf{85.56} \\
    \midrule
    \multirow{6}[2]{*}{|D|=128*$|\mathcal{Y}|$} & BERT  & 77.58 & 69.84 & 83.40 & 80.62 & 78.00 & 77.15 \\
          & RoBERTa & 78.30 & 71.37 & 83.61 & 81.10 & 77.63 & 77.29 \\
          & SciBERT & 82.54 & 75.71 & 83.02 & 80.73 & 84.25 & 84.03 \\
          & PT-soft & 81.46 & \textbf{81.53} & 85.28 & 83.25 & 88.50 & 88.28 \\
          & PT-hard & 80.02 & 73.57 & 86.14 & 84.90 & 88.75 & 88.48 \\
          & MPT   & \textbf{83.79} & 77.89 & \textbf{88.93} & \textbf{87.82} & \textbf{90.38} & \textbf{90.40} \\
    \bottomrule
    \end{tabular}%
    }
  \label{tab:unbal_sup}%
\end{table}%

\begin{table}[!htbp]
  \centering
  \caption{Low-resource experimental results of performance (\%) comparison with semi-supervised baselines on RCT-20k, SciCite, and PMO-kw test sets. Note that the training data are constructed by randomly sampling $K*|\mathcal{Y}|$ instances from the original training set, where $K=\{4,8,16,32,64,128\}$. \textbf{\textit{Bold}} shows the best performance corresponding to $K$.}
  \resizebox{0.48\textwidth}{!}{
    \begin{tabular}{clcccccc}
    \toprule
    \multirow{2}[4]{*}{\# Examples$=K*|\mathcal{Y}|$} & \multicolumn{1}{c}{\multirow{2}[4]{*}{Method}} & \multicolumn{2}{c}{RCT-20k} & \multicolumn{2}{c}{SciCite} & \multicolumn{2}{c}{PMO-kw} \\
\cmidrule{3-8}          &       & Accuracy & Macro F1 & Accuracy & Macro F1 & Accuracy & Macro F1 \\
    \midrule
    \midrule
    \multirow{6}[2]{*}{|D|=4*$|\mathcal{Y}|$} & TMix  & 20.31 & 14.40 & 36.43 & 31.78 & 35.13 & 34.39 \\
          & UDA   & 32.37 & 27.51 & 52.09 & \textbf{45.36} & 36.36 & 17.78 \\
          & MixText & 16.42 & 12.25 & 39.23 & 32.07 & 40.25 & 30.34 \\
          & PET   & 63.68 & 41.42 & 61.58 & 38.72 & \textbf{45.75} & 23.78 \\
          & iPET  & 59.23 & 38.78 & 61.15 & 41.78 & 42.13 & 37.79 \\
          & MPT   & \textbf{73.14} & \textbf{55.04} & \textbf{64.59} & 42.74 & 45.00 & \textbf{42.11} \\
    \midrule
    \multirow{6}[2]{*}{|D|=8*$|\mathcal{Y}|$} & TMix  & 25.79 & 13.96 & 39.01 & 34.91 & 36.13 & 34.83 \\
          & UDA   & 58.08 & 55.88 & 61.58 & \textbf{57.97} & 34.78 & 32.11 \\
          & MixText & 21.95 & 14.97 & 47.45 & 35.82 & 35.50 & 32.05 \\
          & PET   & 70.46 & 49.74 & 56.74 & 30.26 & 46.75 & 43.13 \\
          & iPET  & \textbf{79.85} & \textbf{71.65} & 66.09 & 50.27 & 53.88 & 53.97 \\
          & MPT   & 77.92 & 68.68 & \textbf{77.32} & 55.77 & \textbf{58.87} & \textbf{57.45} \\
    \midrule
    \multirow{6}[2]{*}{|D|=16*$|\mathcal{Y}|$} & TMix  & 35.97 & 17.16 & 45.46 & 34.74 & 41.00 & 35.83 \\
          & UDA   & 65.93 & 65.10 & 70.27 & 69.93 & 40.43 & 36.57 \\
          & MixText & 19.86 & 14.26 & 51.59 & 38.53 & 36.63 & 35.94 \\
          & PET   & 81.98 & 74.69 & 71.31 & 51.11 & 50.13 & 48.13 \\
          & iPET  & 81.23 & 74.15 & 77.27 & 63.79 & 54.63 & 54.24 \\
          & MPT   & \textbf{82.43} & \textbf{75.32} & \textbf{84.09} & \textbf{78.91} & \textbf{68.37} & \textbf{66.50} \\
    \midrule
    \multirow{6}[2]{*}{|D|=32*$|\mathcal{Y}|$} & TMix  & 35.65 & 17.28 & 50.83 & 38.72 & 38.50 & 34.24 \\
          & UDA   & 70.19 & 69.80 & 74.24 & 73.89 & 35.79 & 32.37 \\
          & MixText & 26.38 & 20.34 & 53.63 & 40.25 & 37.88 & 35.52 \\
          & PET   & \textbf{83.52} & \textbf{76.91} & 82.05 & 76.53 & 57.50 & 56.87 \\
          & iPET  & 82.80 & 76.03 & 85.44 & 83.03 & 57.88 & 57.22 \\
          & MPT   & 82.89 & 74.73 & \textbf{87.91} & \textbf{86.55} & \textbf{73.00} & \textbf{72.78} \\
    \midrule
    \multirow{6}[2]{*}{|D|=64*$|\mathcal{Y}|$} & TMix  & 26.99 & 23.03 & 47.18 & 39.01 & 41.88 & 34.68 \\
          & UDA   & 73.98 & 74.39 & 76.36 & 76.30 & 38.74 & 34.23 \\
          & MixText & 32.57 & 22.66 & 52.12 & 45.45 & 39.38 & 35.09 \\
          & PET   & 83.44 & \textbf{76.55} & 87.64 & 86.25 & 60.38 & 59.89 \\
          & iPET  & 82.69 & 75.88 & \textbf{88.66} & \textbf{87.53} & 61.75 & 61.49 \\
          & MPT   & \textbf{83.71} & 75.43 & 88.18 & 86.90 & \textbf{85.62} & \textbf{85.56} \\
    \midrule
    \multirow{6}[2]{*}{|D|=128*$|\mathcal{Y}|$} & TMix  & 35.41 & 16.72 & 51.10 & 32.90 & 42.88 & 34.85 \\
          & UDA   & 76.37 & 76.75 & 80.76 & 80.69 & 44.65 & 20.58 \\
          & MixText & 27.67 & 19.13 & 52.28 & 39.63 & 41.38 & 36.22 \\
          & PET   & 83.53 & 77.25 & 88.34 & 87.32 & 65.63 & 65.69 \\
          & iPET  & 83.18 & 76.99 & 88.23 & 87.28 & 66.63 & 66.67 \\
          & MPT   & \textbf{83.79} & \textbf{77.89} & \textbf{88.93} & \textbf{87.82} & \textbf{90.38} & \textbf{90.40} \\
    \bottomrule
    \end{tabular}%
    }
  \label{tab:unbal_semi}%
\end{table}%

\section{Analysis and discussion}
\label{sec:analyze}

\begin{figure}[!t]
\centering
\includegraphics[width=3.5in]{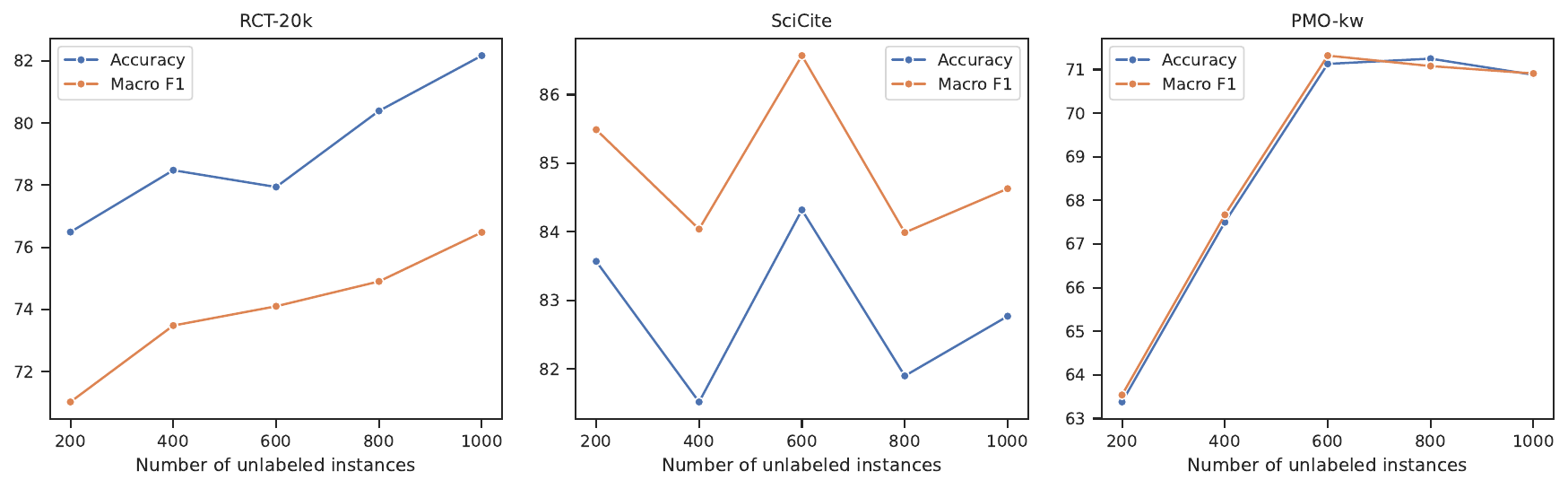}
\caption{The 16-shot performance of MPT using different number of unlabeled instances on RCT-20k, SciCite, and PMO-kw datasets.}
\label{fig_analysis_unlabeled}
\end{figure}

Since we focus on SciBERT based MPT using 600 or 1000 unlabeled instances in the main experiments, we further conduct experiments to better understand the effect of different amount of unlabeled instances, different backbone PLMs, and in-domain pre-train.

\subsection{Effect of different number of unlabeled instances}
\label{unlabeled_samples}
We also conduct experiments, which are shown in Figure \ref{fig_analysis_unlabeled}, to test our model performances with 16 instances per class and different amount of unlabeled data (from 200 to 1000) on Rct-20k, SciCite, and PMO-kw datasets. We can observe that, although for different datasets, models vary in the choice of optimal unlabeled instance size. Overall, with more unlabeled data, the overall performance of MPT
becomes much higher. It validates the effectiveness of our proposed MPT in making full use of unlabeled data.

\subsection{Can MPT applied to other PLMs?}
\begin{figure}[!t]
\centering
\includegraphics[width=3.5in]{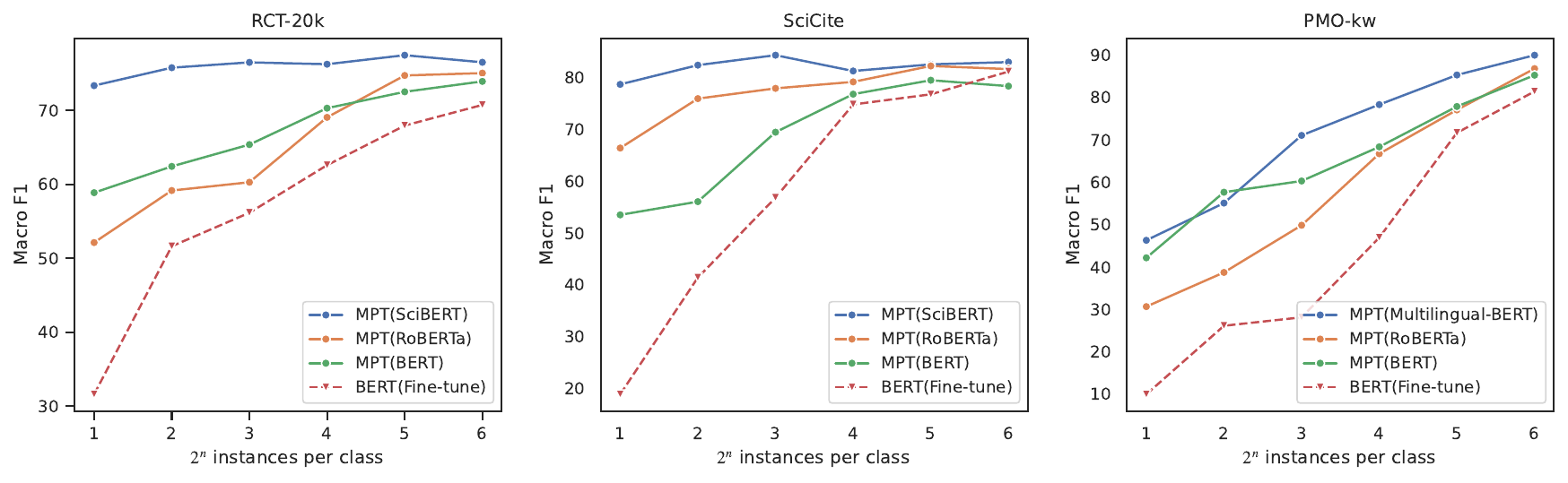}
\caption{The few-shot performance of MPT on RCT-20k, SciCite, and PMO-kw datasets utilizing different PLMs as backbone models. The performances of BERT fine-tuning are selected as referential performances.}
\label{fig_analysis_plm}
\end{figure}
In this paper, we focus on adopting SciBERT as the backbone of MPT in the main experiments. Can we further extend MPT to other PLMs like BERT and RoBERTa with different pre-train corpus and strategy? To achieve this, we replace the SciBERT or Multilingual BERT in MPT into BERT and RoBERTa. Meanwhile, we keep the same prompt templates and verbalizers with the SciBERT based MPT. As shown in Figure \ref{fig_analysis_plm}, it depicts that MPT performs significantly better than fine-tuning when data is extremely scarce. As the volume of data grows, BERT fine-tuning shows comparative performance with BERT based MPT. Moreover, part of the performance improvement of RoBERT or SciBERT based MPT over BERT based MPT may be due to the contribution of pre-train knowledge. We will further discuss it in the next subsection.

\subsection{Effect of in-domain pre-training}
\begin{figure}[!t]
\centering
\includegraphics[width=3.5in]{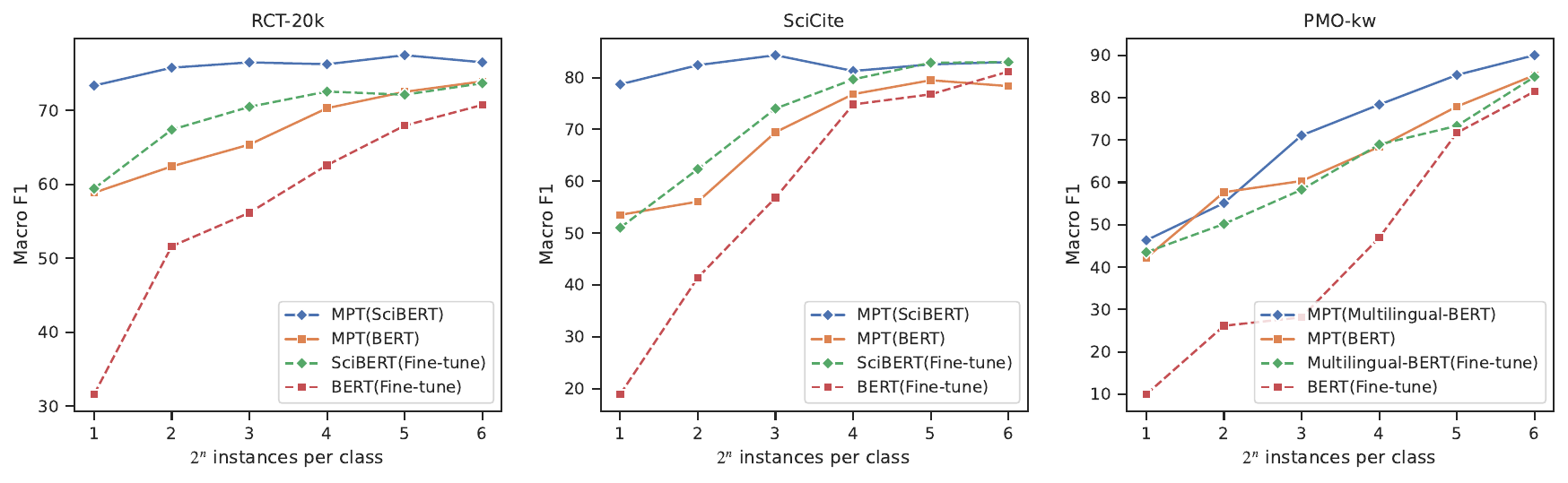}
\caption{The performance of BERT fine-tuning and MPT both with (BERT) and without (SciBERT or multilingual BERT) pre-training in scientific domain or multilingual-domain.}
\label{fig_analysis_further}
\end{figure}
Compared to supervised baselines, our proposed MPT and other semi-supervised baselines utilize SciBERT or multilingual BERT. As a result, part of the performance improvement might come from the additional in-domain pre-train corpus. Thus, we compare BERT and SciBERT/Multilingual-BERT based MPT with BERT and SciBERT fine-tuning to test the effect of in-domain pre-training. SciBERT utilizes the same structure as BERT but is pre-trained on scientific domain data, as in-domain pre-train is a common way to improve the model performance \citep{gururangan2020don,sung2019pre}. As shown in Figure \ref{fig_analysis_further} shows results of fine-tuning and MPT both using BERT and SciBERT. We can observe that, compared with BERT based MPT, although SciBERT fine-tuning achieves significant better performance when data is extremely scarce and comparative performance with the volume of data grows, it still performs worse than MPT. This observation indicates that MPT not only makes full use of the unlabeled data but also stimulates the pre-train knowledge effectively.

\subsection{Comparison with Large Language Models}
\begin{table}[!t]
  \centering
  \caption{Few-shot experimental results of performance (\%) comparison with widely used large language models on RCT-20k and SciCite test sets.}
    \begin{tabular}{lrr}
    \toprule
    \multicolumn{1}{c}{Method} & \multicolumn{1}{c}{RCT-20k} & \multicolumn{1}{c}{SciCite} \\
    \midrule
    MPT   & \textbf{73.34} & \textbf{78.69} \\
    GPT-4 & 72.74 & 43.99 \\
    ERNIE-bot & 67.08 & 43.64 \\
    ChatGLM2 & 58.43 & 35.77 \\
    GPT-3.5-turbo & 64.19 & 26.34 \\
    \bottomrule
    \end{tabular}%
  \label{tab:llm}%
\end{table}%

The rapid development of large language models (LLMs), exemplified by ChatGPT, has attracted extensive attention. These massively PLMs, which undergo instruction fine-tuning and align with human intent, have demonstrated excellent performance not only in generative tasks, such as question answering and machine translation, but also in diverse tasks, such as information extraction, text classification, and reasoning when guided by prompts. Therefore, we further compared the 4-shot performance of our proposed MPT with the best 3 to 5 shot best performance of mainstream LLMs, e.g., GPT-4, GPT-3.5-turbo, ERNIE-bot, and ChatGLM2, in the few-shot scenario. Consistent with previous works \citep{zhang2022active,dong2022survey,xu2024misconfidence,liu2024let}, we evaluate the performance in an In-Context Learning manner. As shown in Table \ref{tab:llm}, MPT significantly outperforms these models.

\subsection{Connection and comparison with existing work}
Our proposed MPT can achieve superior performance for citation function recognition, structure function recognition, and keyword function recognition in different low-resource settings. Previous methods on academic function recognition tasks can mainly be summarized in three categories: (1) a machine learning method based on hand-crafted features \citep{jurgens2016citation,lu2018functional,nanba2010automatic}; (2) a deep learning model trained from scratch \citep{cohan2019structural,qin2020using,qian2020structure,cheng2021recognition}; (3) a PLM fine-tuning method \citep{beltagy2019scibert,huang2022disclosing,wei2020recognition,zhang2021multiword}. Those methods require a large number of annotated data to achieve competitive performance. Our proposed method is based on prompt learning, which is a new paradigm of utilizing PLMs \citep{liu2021pre} and has great potential in low-resource scenarios. It utilize prompts to bridge the gap of objective forms in pre-training and fine-tuning, which leads to more effective utilization of pre-train knowledge than the standard fine-tuning. Moreover, we introduce prompt learning with mixing multiple types of prompt templates. Whereas previous s studies in other domain tasks are solely based on manual hard templates or automatically learned soft templates. Our comparative experiment results between fine-tuning and prompt learning in Section \ref{results} further validate its effectiveness for low-resource academic function recognition tasks.

Our proposed MPT is a semi-supervised method that utilizes multiple prompt templates to annotate unlabeled data. The semi-supervised framework is similar to iPET \citep{schick2021exploiting}, one of the strong semi-supervised baselines. Differently, compared with this method that only adopts the manual prompt, we combine the manual prompt with automatically learned continuous prompt, which can provide multi-perspective representations and take full advantage of knowledge in the PLM and unlabeled data.

Moreover, existing works lack attention to scientific classification tasks in low-resource settings. Since obtaining the annotated data for scientific NLP tasks is still challenging and expensive, it is common in real-world scenarios that there is usually no annotated data or only a small number of annotated instances. To alleviate the dependence on annotated data for scientific classification tasks, we propose the MPT, which combines multiple prompts with a semi-supervised framework. Extensive experiments on a series of academic function recognition tasks at different granularities prove the feasibility of MPT.

\subsection{Implications for research}
This study has the following implications. First, in the practical scientific NLP scenario, there is a contradiction between the massive unlabeled scientific publications and the scarcity of annotated data. To this end, we propose MPT, a semi-supervised and prompt learning based solution coping with practical low-resource academic function recognition scenarios. Second, the prompt learning paradigm is promising for low-resource scientific NLP tasks. Moreover, MPT is a semi-supervised solution and fuses the manual prompt with automatically learned continuous prompt. It provides multi-perspective representations and takes full advantage of knowledge in the PLM and unlabeled data resources. Third, our proposed MPT has the ability to perform multi-granularity academic function recognition. Moreover, the MPT presented in this study is a general approach that can be easily deployed in other scientific NLP tasks with minor adjustment to the prompt templates and verbalizer. Finally, MPT is a method for low-resource scenario which can be considered a type of \textit{Green AI} approach, as they aim to develop and use models in a way that is more resource-efficient. By using the method, it is possible to build and train models with less data and compute power, which can reduce the environmental impact of the AI system.

\section{Conclusion}
In this paper, we propose several prompts and introduce prompt learning method for different granularity academic function recognition tasks. Then we present Mix Prompt Tuning (MPT), a semi-supervised solution that combines the manual prompt with automatically learned continuous prompt for different granularity academic function recognition tasks in practical scenario. Extensive experiments demonstrate that our proposed method outperforms other fine-tuning, prompt-tuning , or semi-supervised baselines.

There are several important directions for future work: (1) inject latent knowledge contained in knowledge graph and citation graph into prompt construction and tuning to increase the interpretability and further alleviate the dependence on the manual prompt. (2) make full use of prompt to exploit pre-trained language models for better scientific fact prediction. (3) investigate the transferability of prompt tuning across different scientific tasks and models.

\section*{Acknowledgments}
This work is supported by the National Natural Science Foundation of China (72234005 and 72174157).

\bibliography{sample-base}


\begin{thebibliography}{79}


\ifx \showCODEN    \undefined \def \showCODEN     #1{\unskip}     \fi
\ifx \showDOI      \undefined \def \showDOI       #1{#1}\fi
\ifx \showISBNx    \undefined \def \showISBNx     #1{\unskip}     \fi
\ifx \showISBNxiii \undefined \def \showISBNxiii  #1{\unskip}     \fi
\ifx \showISSN     \undefined \def \showISSN      #1{\unskip}     \fi
\ifx \showLCCN     \undefined \def \showLCCN      #1{\unskip}     \fi
\ifx \shownote     \undefined \def \shownote      #1{#1}          \fi
\ifx \showarticletitle \undefined \def \showarticletitle #1{#1}   \fi
\ifx \showURL      \undefined \def \showURL       {\relax}        \fi
\providecommand\bibfield[2]{#2}
\providecommand\bibinfo[2]{#2}
\providecommand\natexlab[1]{#1}
\providecommand\showeprint[2][]{arXiv:#2}

\bibitem[Bao et~al\mbox{.}(2019)]%
        {bao2019few}
\bibfield{author}{\bibinfo{person}{Yujia Bao}, \bibinfo{person}{Menghua Wu},
  \bibinfo{person}{Shiyu Chang}, {and} \bibinfo{person}{Regina Barzilay}.}
  \bibinfo{year}{2019}\natexlab{}.
\newblock \showarticletitle{Few-shot Text Classification with Distributional
  Signatures}. In \bibinfo{booktitle}{\emph{International Conference on
  Learning Representations}}.
\newblock


\bibitem[Beltagy et~al\mbox{.}(2019)]%
        {beltagy2019scibert}
\bibfield{author}{\bibinfo{person}{Iz Beltagy}, \bibinfo{person}{Kyle Lo},
  {and} \bibinfo{person}{Arman Cohan}.} \bibinfo{year}{2019}\natexlab{}.
\newblock \showarticletitle{SciBERT: A pretrained language model for scientific
  text}.
\newblock \bibinfo{journal}{\emph{arXiv preprint arXiv:1903.10676}}
  (\bibinfo{year}{2019}).
\newblock


\bibitem[Brown et~al\mbox{.}(2020)]%
        {brown2020language}
\bibfield{author}{\bibinfo{person}{Tom Brown}, \bibinfo{person}{Benjamin Mann},
  \bibinfo{person}{Nick Ryder}, \bibinfo{person}{Melanie Subbiah},
  \bibinfo{person}{Jared~D Kaplan}, \bibinfo{person}{Prafulla Dhariwal},
  \bibinfo{person}{Arvind Neelakantan}, \bibinfo{person}{Pranav Shyam},
  \bibinfo{person}{Girish Sastry}, \bibinfo{person}{Amanda Askell},
  {et~al\mbox{.}}} \bibinfo{year}{2020}\natexlab{}.
\newblock \showarticletitle{Language models are few-shot learners}.
\newblock \bibinfo{journal}{\emph{Advances in neural information processing
  systems}}  \bibinfo{volume}{33} (\bibinfo{year}{2020}),
  \bibinfo{pages}{1877--1901}.
\newblock


\bibitem[Chen et~al\mbox{.}(2020a)]%
        {chen2020semi}
\bibfield{author}{\bibinfo{person}{Jiaao Chen}, \bibinfo{person}{Yuwei Wu},
  {and} \bibinfo{person}{Diyi Yang}.} \bibinfo{year}{2020}\natexlab{a}.
\newblock \showarticletitle{Semi-supervised models via data augmentationfor
  classifying interactive affective responses}.
\newblock \bibinfo{journal}{\emph{arXiv preprint arXiv:2004.10972}}
  (\bibinfo{year}{2020}).
\newblock


\bibitem[Chen et~al\mbox{.}(2020b)]%
        {chen2020mixtext}
\bibfield{author}{\bibinfo{person}{Jiaao Chen}, \bibinfo{person}{Zichao Yang},
  {and} \bibinfo{person}{Diyi Yang}.} \bibinfo{year}{2020}\natexlab{b}.
\newblock \showarticletitle{MixText: Linguistically-Informed Interpolation of
  Hidden Space for Semi-Supervised Text Classification}. In
  \bibinfo{booktitle}{\emph{Proceedings of the 58th Annual Meeting of the
  Association for Computational Linguistics}}. \bibinfo{pages}{2147--2157}.
\newblock


\bibitem[Cheng et~al\mbox{.}(2021)]%
        {cheng2021recognition}
\bibfield{author}{\bibinfo{person}{Qikai Cheng}, \bibinfo{person}{Pengcheng
  Li}, \bibinfo{person}{Guobiao Zhang}, {and} \bibinfo{person}{Wei Lu}.}
  \bibinfo{year}{2021}\natexlab{}.
\newblock \showarticletitle{Recognition of Lexical Functions in Academic Texts:
  Problem Method Extraction Based on Title Generation Strategy and Attention
  Mechanism}.
\newblock \bibinfo{journal}{\emph{Journal of the China Society for Scientific
  and Technical Information}} \bibinfo{volume}{40}, \bibinfo{number}{1},
  Article \bibinfo{articleno}{43} (\bibinfo{year}{2021}),
  \bibinfo{numpages}{9}~pages.
\newblock


\bibitem[Cohan et~al\mbox{.}(2019)]%
        {cohan2019structural}
\bibfield{author}{\bibinfo{person}{Arman Cohan}, \bibinfo{person}{Waleed
  Ammar}, \bibinfo{person}{Madeleine Van~Zuylen}, {and} \bibinfo{person}{Field
  Cady}.} \bibinfo{year}{2019}\natexlab{}.
\newblock \showarticletitle{Structural scaffolds for citation intent
  classification in scientific publications}.
\newblock \bibinfo{journal}{\emph{arXiv preprint arXiv:1904.01608}}
  (\bibinfo{year}{2019}).
\newblock


\bibitem[Cui et~al\mbox{.}(2022)]%
        {cui2022prototypical}
\bibfield{author}{\bibinfo{person}{Ganqu Cui}, \bibinfo{person}{Shengding Hu},
  \bibinfo{person}{Ning Ding}, \bibinfo{person}{Longtao Huang}, {and}
  \bibinfo{person}{Zhiyuan Liu}.} \bibinfo{year}{2022}\natexlab{}.
\newblock \showarticletitle{Prototypical Verbalizer for Prompt-based Few-shot
  Tuning}. In \bibinfo{booktitle}{\emph{Proceedings of the 60th Annual Meeting
  of the Association for Computational Linguistics (Volume 1: Long Papers)}}.
  \bibinfo{pages}{7014--7024}.
\newblock


\bibitem[Dernoncourt and Lee(2017)]%
        {dernoncourt2017pubmed}
\bibfield{author}{\bibinfo{person}{Franck Dernoncourt} {and}
  \bibinfo{person}{Ji~Young Lee}.} \bibinfo{year}{2017}\natexlab{}.
\newblock \showarticletitle{PubMed 200k RCT: a Dataset for Sequential Sentence
  Classification in Medical Abstracts}. In
  \bibinfo{booktitle}{\emph{Proceedings of the Eighth International Joint
  Conference on Natural Language Processing (Volume 2: Short Papers)}}.
  \bibinfo{pages}{308--313}.
\newblock


\bibitem[Dernoncourt et~al\mbox{.}(2016)]%
        {dernoncourt2016neural}
\bibfield{author}{\bibinfo{person}{Franck Dernoncourt},
  \bibinfo{person}{Ji~Young Lee}, {and} \bibinfo{person}{Peter Szolovits}.}
  \bibinfo{year}{2016}\natexlab{}.
\newblock \showarticletitle{Neural networks for joint sentence classification
  in medical paper abstracts}.
\newblock \bibinfo{journal}{\emph{arXiv preprint arXiv:1612.05251}}
  (\bibinfo{year}{2016}).
\newblock


\bibitem[Devlin et~al\mbox{.}(2019)]%
        {devlin2019bert}
\bibfield{author}{\bibinfo{person}{Jacob Devlin}, \bibinfo{person}{Ming-Wei
  Chang}, \bibinfo{person}{Kenton Lee}, {and} \bibinfo{person}{Kristina
  Toutanova}.} \bibinfo{year}{2019}\natexlab{}.
\newblock \showarticletitle{BERT: Pre-training of Deep Bidirectional
  Transformers for Language Understanding}. In
  \bibinfo{booktitle}{\emph{Proceedings of NAACL-HLT}}.
  \bibinfo{pages}{4171--4186}.
\newblock


\bibitem[Ding et~al\mbox{.}(2021)]%
        {ding2021openprompt}
\bibfield{author}{\bibinfo{person}{Ning Ding}, \bibinfo{person}{Shengding Hu},
  \bibinfo{person}{Weilin Zhao}, \bibinfo{person}{Yulin Chen},
  \bibinfo{person}{Zhiyuan Liu}, \bibinfo{person}{Hai-Tao Zheng}, {and}
  \bibinfo{person}{Maosong Sun}.} \bibinfo{year}{2021}\natexlab{}.
\newblock \showarticletitle{Openprompt: An open-source framework for
  prompt-learning}.
\newblock \bibinfo{journal}{\emph{arXiv preprint arXiv:2111.01998}}
  (\bibinfo{year}{2021}).
\newblock


\bibitem[Dong et~al\mbox{.}(2022)]%
        {dong2022survey}
\bibfield{author}{\bibinfo{person}{Qingxiu Dong}, \bibinfo{person}{Lei Li},
  \bibinfo{person}{Damai Dai}, \bibinfo{person}{Ce Zheng},
  \bibinfo{person}{Zhiyong Wu}, \bibinfo{person}{Baobao Chang},
  \bibinfo{person}{Xu Sun}, \bibinfo{person}{Jingjing Xu}, {and}
  \bibinfo{person}{Zhifang Sui}.} \bibinfo{year}{2022}\natexlab{}.
\newblock \showarticletitle{A survey on in-context learning}.
\newblock \bibinfo{journal}{\emph{arXiv preprint arXiv:2301.00234}}
  (\bibinfo{year}{2022}).
\newblock


\bibitem[Gao et~al\mbox{.}(2021)]%
        {gao2021making}
\bibfield{author}{\bibinfo{person}{Tianyu Gao}, \bibinfo{person}{Adam Fisch},
  {and} \bibinfo{person}{Danqi Chen}.} \bibinfo{year}{2021}\natexlab{}.
\newblock \showarticletitle{Making Pre-trained Language Models Better Few-shot
  Learners}. In \bibinfo{booktitle}{\emph{Proceedings of the 59th Annual
  Meeting of the Association for Computational Linguistics and the 11th
  International Joint Conference on Natural Language Processing (Volume 1: Long
  Papers)}}. \bibinfo{pages}{3816--3830}.
\newblock


\bibitem[Guo et~al\mbox{.}(2017)]%
        {guo2017calibration}
\bibfield{author}{\bibinfo{person}{Chuan Guo}, \bibinfo{person}{Geoff Pleiss},
  \bibinfo{person}{Yu Sun}, {and} \bibinfo{person}{Kilian~Q Weinberger}.}
  \bibinfo{year}{2017}\natexlab{}.
\newblock \showarticletitle{On calibration of modern neural networks}. In
  \bibinfo{booktitle}{\emph{International Conference on Machine Learning}}.
  PMLR, \bibinfo{pages}{1321--1330}.
\newblock


\bibitem[Gururangan et~al\mbox{.}(2020)]%
        {gururangan2020don}
\bibfield{author}{\bibinfo{person}{Suchin Gururangan}, \bibinfo{person}{Ana
  Marasovi{\'c}}, \bibinfo{person}{Swabha Swayamdipta}, \bibinfo{person}{Kyle
  Lo}, \bibinfo{person}{Iz Beltagy}, \bibinfo{person}{Doug Downey}, {and}
  \bibinfo{person}{Noah~A Smith}.} \bibinfo{year}{2020}\natexlab{}.
\newblock \showarticletitle{Don’t Stop Pretraining: Adapt Language Models to
  Domains and Tasks}. In \bibinfo{booktitle}{\emph{Proceedings of the 58th
  Annual Meeting of the Association for Computational Linguistics}}.
  \bibinfo{pages}{8342--8360}.
\newblock


\bibitem[Hambardzumyan et~al\mbox{.}(2021)]%
        {hambardzumyan2021warp}
\bibfield{author}{\bibinfo{person}{Karen Hambardzumyan}, \bibinfo{person}{Hrant
  Khachatrian}, {and} \bibinfo{person}{Jonathan May}.}
  \bibinfo{year}{2021}\natexlab{}.
\newblock \showarticletitle{WARP: Word-level Adversarial ReProgramming}. In
  \bibinfo{booktitle}{\emph{Proceedings of the 59th Annual Meeting of the
  Association for Computational Linguistics and the 11th International Joint
  Conference on Natural Language Processing (Volume 1: Long Papers)}}.
  \bibinfo{pages}{4921--4933}.
\newblock


\bibitem[Han et~al\mbox{.}(2021a)]%
        {han2021pre}
\bibfield{author}{\bibinfo{person}{Xu Han}, \bibinfo{person}{Zhengyan Zhang},
  \bibinfo{person}{Ning Ding}, \bibinfo{person}{Yuxian Gu},
  \bibinfo{person}{Xiao Liu}, \bibinfo{person}{Yuqi Huo},
  \bibinfo{person}{Jiezhong Qiu}, \bibinfo{person}{Yuan Yao},
  \bibinfo{person}{Ao Zhang}, \bibinfo{person}{Liang Zhang}, {et~al\mbox{.}}}
  \bibinfo{year}{2021}\natexlab{a}.
\newblock \showarticletitle{Pre-trained models: Past, present and future}.
\newblock \bibinfo{journal}{\emph{AI Open}}  \bibinfo{volume}{2}
  (\bibinfo{year}{2021}), \bibinfo{pages}{225--250}.
\newblock


\bibitem[Han et~al\mbox{.}(2021b)]%
        {han2021ptr}
\bibfield{author}{\bibinfo{person}{Xu Han}, \bibinfo{person}{Weilin Zhao},
  \bibinfo{person}{Ning Ding}, \bibinfo{person}{Zhiyuan Liu}, {and}
  \bibinfo{person}{Maosong Sun}.} \bibinfo{year}{2021}\natexlab{b}.
\newblock \showarticletitle{Ptr: Prompt tuning with rules for text
  classification}.
\newblock \bibinfo{journal}{\emph{arXiv preprint arXiv:2105.11259}}
  (\bibinfo{year}{2021}).
\newblock


\bibitem[Hassanzadeh et~al\mbox{.}(2014)]%
        {hassanzadeh2014identifying}
\bibfield{author}{\bibinfo{person}{Hamed Hassanzadeh}, \bibinfo{person}{Tudor
  Groza}, {and} \bibinfo{person}{Jane Hunter}.}
  \bibinfo{year}{2014}\natexlab{}.
\newblock \showarticletitle{Identifying scientific artefacts in biomedical
  literature: The evidence based medicine use case}.
\newblock \bibinfo{journal}{\emph{Journal of biomedical informatics}}
  \bibinfo{volume}{49} (\bibinfo{year}{2014}), \bibinfo{pages}{159--170}.
\newblock


\bibitem[Hewitt and Manning(2019)]%
        {hewitt2019structural}
\bibfield{author}{\bibinfo{person}{John Hewitt} {and}
  \bibinfo{person}{Christopher~D Manning}.} \bibinfo{year}{2019}\natexlab{}.
\newblock \showarticletitle{A structural probe for finding syntax in word
  representations}. In \bibinfo{booktitle}{\emph{Proceedings of the 2019
  Conference of the North American Chapter of the Association for Computational
  Linguistics: Human Language Technologies, Volume 1 (Long and Short Papers)}}.
  \bibinfo{pages}{4129--4138}.
\newblock


\bibitem[Hu et~al\mbox{.}(2021)]%
        {hu2021knowledgeable}
\bibfield{author}{\bibinfo{person}{Shengding Hu}, \bibinfo{person}{Ning Ding},
  \bibinfo{person}{Huadong Wang}, \bibinfo{person}{Zhiyuan Liu},
  \bibinfo{person}{Juanzi Li}, {and} \bibinfo{person}{Maosong Sun}.}
  \bibinfo{year}{2021}\natexlab{}.
\newblock \showarticletitle{Knowledgeable prompt-tuning: Incorporating
  knowledge into prompt verbalizer for text classification}.
\newblock \bibinfo{journal}{\emph{arXiv preprint arXiv:2108.02035}}
  (\bibinfo{year}{2021}).
\newblock


\bibitem[Huang et~al\mbox{.}(2022)]%
        {huang2022disclosing}
\bibfield{author}{\bibinfo{person}{Shengzhi Huang}, \bibinfo{person}{Jiajia
  Qian}, \bibinfo{person}{Yong Huang}, \bibinfo{person}{Wei Lu},
  \bibinfo{person}{Yi Bu}, \bibinfo{person}{Jinqing Yang}, {and}
  \bibinfo{person}{Qikai Cheng}.} \bibinfo{year}{2022}\natexlab{}.
\newblock \showarticletitle{Disclosing the relationship between citation
  structure and future impact of a publication}.
\newblock \bibinfo{journal}{\emph{Journal of the Association for Information
  Science and Technology}} \bibinfo{volume}{73}, \bibinfo{number}{7}
  (\bibinfo{year}{2022}), \bibinfo{pages}{1025--1042}.
\newblock


\bibitem[Huang et~al\mbox{.}(2016a)]%
        {huang2016sec}
\bibfield{author}{\bibinfo{person}{Yong Huang}, \bibinfo{person}{Wei Lu}, {and}
  \bibinfo{person}{Qikai Cheng}.} \bibinfo{year}{2016}\natexlab{a}.
\newblock \showarticletitle{The Structure Function Recognition of Academic
  Text——Chapter Content Based Recognition}.
\newblock \bibinfo{journal}{\emph{Journal of the China Society for Scientific
  and Technical Information}} \bibinfo{volume}{35}, \bibinfo{number}{03}
  (\bibinfo{year}{2016}), \bibinfo{pages}{293--300}.
\newblock


\bibitem[Huang et~al\mbox{.}(2016b)]%
        {huang2016para}
\bibfield{author}{\bibinfo{person}{Yong Huang}, \bibinfo{person}{Wei Lu},
  \bibinfo{person}{Qikai Cheng}, {and} \bibinfo{person}{Sisi Gui}.}
  \bibinfo{year}{2016}\natexlab{b}.
\newblock \showarticletitle{The Structure Function Recognition of Academic
  Text——Paragraph-based Recognition}.
\newblock \bibinfo{journal}{\emph{Journal of the China Society for Scientific
  and Technical Information}} \bibinfo{volume}{35}, \bibinfo{number}{05}
  (\bibinfo{year}{2016}), \bibinfo{pages}{530--538}.
\newblock


\bibitem[Huo et~al\mbox{.}(2022)]%
        {huo2022hotness}
\bibfield{author}{\bibinfo{person}{Chaoguang Huo}, \bibinfo{person}{Shutian
  Ma}, {and} \bibinfo{person}{Xiaozhong Liu}.} \bibinfo{year}{2022}\natexlab{}.
\newblock \showarticletitle{Hotness prediction of scientific topics based on a
  bibliographic knowledge graph}.
\newblock \bibinfo{journal}{\emph{Information Processing \& Management}}
  \bibinfo{volume}{59}, \bibinfo{number}{4} (\bibinfo{year}{2022}),
  \bibinfo{pages}{102980}.
\newblock


\bibitem[Jawahar et~al\mbox{.}(2019)]%
        {jawahar2019does}
\bibfield{author}{\bibinfo{person}{Ganesh Jawahar},
  \bibinfo{person}{Beno{\^\i}t Sagot}, {and} \bibinfo{person}{Djam{\'e}
  Seddah}.} \bibinfo{year}{2019}\natexlab{}.
\newblock \showarticletitle{What Does BERT Learn about the Structure of
  Language?}. In \bibinfo{booktitle}{\emph{Proceedings of the 57th Annual
  Meeting of the Association for Computational Linguistics}}.
  \bibinfo{pages}{3651--3657}.
\newblock


\bibitem[Jiang et~al\mbox{.}(2022)]%
        {jiang2022fake}
\bibfield{author}{\bibinfo{person}{Gongyao Jiang}, \bibinfo{person}{Shuang
  Liu}, \bibinfo{person}{Yu Zhao}, \bibinfo{person}{Yueheng Sun}, {and}
  \bibinfo{person}{Meishan Zhang}.} \bibinfo{year}{2022}\natexlab{}.
\newblock \showarticletitle{Fake news detection via knowledgeable prompt
  learning}.
\newblock \bibinfo{journal}{\emph{Information Processing \& Management}}
  \bibinfo{volume}{59}, \bibinfo{number}{5} (\bibinfo{year}{2022}),
  \bibinfo{pages}{103029}.
\newblock


\bibitem[Jin and Szolovits(2018)]%
        {jin2018hierarchical}
\bibfield{author}{\bibinfo{person}{Di Jin} {and} \bibinfo{person}{Peter
  Szolovits}.} \bibinfo{year}{2018}\natexlab{}.
\newblock \showarticletitle{Hierarchical neural networks for sequential
  sentence classification in medical scientific abstracts}.
\newblock \bibinfo{journal}{\emph{arXiv preprint arXiv:1808.06161}}
  (\bibinfo{year}{2018}).
\newblock


\bibitem[Jurgens et~al\mbox{.}(2016)]%
        {jurgens2016citation}
\bibfield{author}{\bibinfo{person}{David Jurgens}, \bibinfo{person}{Srijan
  Kumar}, \bibinfo{person}{Raine Hoover}, \bibinfo{person}{Dan McFarland},
  {and} \bibinfo{person}{Dan Jurafsky}.} \bibinfo{year}{2016}\natexlab{}.
\newblock \showarticletitle{Citation classification for behavioral analysis of
  a scientific field}.
\newblock \bibinfo{journal}{\emph{arXiv preprint arXiv:1609.00435}}
  (\bibinfo{year}{2016}).
\newblock


\bibitem[Kondo et~al\mbox{.}(2009)]%
        {kondo2009technical}
\bibfield{author}{\bibinfo{person}{Tomoki Kondo}, \bibinfo{person}{Hidetsugu
  Nanba}, \bibinfo{person}{Toshiyuki Takezawa}, {and} \bibinfo{person}{Manabu
  Okumura}.} \bibinfo{year}{2009}\natexlab{}.
\newblock \showarticletitle{Technical trend analysis by analyzing research
  papers’ titles}. In \bibinfo{booktitle}{\emph{Language and Technology
  Conference}}. Springer, \bibinfo{pages}{512--521}.
\newblock


\bibitem[Lahav et~al\mbox{.}(2021)]%
        {lahav2021search}
\bibfield{author}{\bibinfo{person}{Dan Lahav}, \bibinfo{person}{Jon~Saad
  Falcon}, \bibinfo{person}{Bailey Kuehl}, \bibinfo{person}{Sophie Johnson},
  \bibinfo{person}{Sravanthi Parasa}, \bibinfo{person}{Noam Shomron},
  \bibinfo{person}{Duen~Horng Chau}, \bibinfo{person}{Diyi Yang},
  \bibinfo{person}{Eric Horvitz}, \bibinfo{person}{Daniel~S Weld},
  {et~al\mbox{.}}} \bibinfo{year}{2021}\natexlab{}.
\newblock \showarticletitle{A Search Engine for Discovery of Scientific
  Challenges and Directions}.
\newblock \bibinfo{journal}{\emph{arXiv preprint arXiv:2108.13751}}
  (\bibinfo{year}{2021}).
\newblock


\bibitem[Lee et~al\mbox{.}(2020)]%
        {lee2020biobert}
\bibfield{author}{\bibinfo{person}{Jinhyuk Lee}, \bibinfo{person}{Wonjin Yoon},
  \bibinfo{person}{Sungdong Kim}, \bibinfo{person}{Donghyeon Kim},
  \bibinfo{person}{Sunkyu Kim}, \bibinfo{person}{Chan~Ho So}, {and}
  \bibinfo{person}{Jaewoo Kang}.} \bibinfo{year}{2020}\natexlab{}.
\newblock \showarticletitle{BioBERT: a pre-trained biomedical language
  representation model for biomedical text mining}.
\newblock \bibinfo{journal}{\emph{Bioinformatics}} \bibinfo{volume}{36},
  \bibinfo{number}{4} (\bibinfo{year}{2020}), \bibinfo{pages}{1234--1240}.
\newblock


\bibitem[Lester et~al\mbox{.}(2021)]%
        {lester2021power}
\bibfield{author}{\bibinfo{person}{Brian Lester}, \bibinfo{person}{Rami
  Al-Rfou}, {and} \bibinfo{person}{Noah Constant}.}
  \bibinfo{year}{2021}\natexlab{}.
\newblock \showarticletitle{The Power of Scale for Parameter-Efficient Prompt
  Tuning}. In \bibinfo{booktitle}{\emph{Proceedings of the 2021 Conference on
  Empirical Methods in Natural Language Processing}}.
  \bibinfo{pages}{3045--3059}.
\newblock


\bibitem[Li and Liang(2021)]%
        {li2021prefix}
\bibfield{author}{\bibinfo{person}{Xiang~Lisa Li} {and} \bibinfo{person}{Percy
  Liang}.} \bibinfo{year}{2021}\natexlab{}.
\newblock \showarticletitle{Prefix-Tuning: Optimizing Continuous Prompts for
  Generation}. In \bibinfo{booktitle}{\emph{Proceedings of the 59th Annual
  Meeting of the Association for Computational Linguistics and the 11th
  International Joint Conference on Natural Language Processing (Volume 1: Long
  Papers)}}. \bibinfo{pages}{4582--4597}.
\newblock


\bibitem[Liang et~al\mbox{.}(2021)]%
        {liang2021combining}
\bibfield{author}{\bibinfo{person}{Zhentao Liang}, \bibinfo{person}{Jin Mao},
  \bibinfo{person}{Kun Lu}, \bibinfo{person}{Zhichao Ba}, {and}
  \bibinfo{person}{Gang Li}.} \bibinfo{year}{2021}\natexlab{}.
\newblock \showarticletitle{Combining deep neural network and bibliometric
  indicator for emerging research topic prediction}.
\newblock \bibinfo{journal}{\emph{Information Processing \& Management}}
  \bibinfo{volume}{58}, \bibinfo{number}{5} (\bibinfo{year}{2021}),
  \bibinfo{pages}{102611}.
\newblock


\bibitem[Liu et~al\mbox{.}(2021a)]%
        {liu2021pre}
\bibfield{author}{\bibinfo{person}{Pengfei Liu}, \bibinfo{person}{Weizhe Yuan},
  \bibinfo{person}{Jinlan Fu}, \bibinfo{person}{Zhengbao Jiang},
  \bibinfo{person}{Hiroaki Hayashi}, {and} \bibinfo{person}{Graham Neubig}.}
  \bibinfo{year}{2021}\natexlab{a}.
\newblock \showarticletitle{Pre-train, prompt, and predict: A systematic survey
  of prompting methods in natural language processing}.
\newblock \bibinfo{journal}{\emph{arXiv preprint arXiv:2107.13586}}
  (\bibinfo{year}{2021}).
\newblock


\bibitem[Liu et~al\mbox{.}(2021b)]%
        {liu2021gpt}
\bibfield{author}{\bibinfo{person}{Xiao Liu}, \bibinfo{person}{Yanan Zheng},
  \bibinfo{person}{Zhengxiao Du}, \bibinfo{person}{Ming Ding},
  \bibinfo{person}{Yujie Qian}, \bibinfo{person}{Zhilin Yang}, {and}
  \bibinfo{person}{Jie Tang}.} \bibinfo{year}{2021}\natexlab{b}.
\newblock \showarticletitle{GPT understands, too}.
\newblock \bibinfo{journal}{\emph{arXiv preprint arXiv:2103.10385}}
  (\bibinfo{year}{2021}).
\newblock


\bibitem[Liu et~al\mbox{.}(2024)]%
        {liu2024let}
\bibfield{author}{\bibinfo{person}{Yinpeng Liu}, \bibinfo{person}{Jiawei Liu},
  \bibinfo{person}{Xiang Shi}, \bibinfo{person}{Qikai Cheng}, {and}
  \bibinfo{person}{Wei Lu}.} \bibinfo{year}{2024}\natexlab{}.
\newblock \showarticletitle{Let's Learn Step by Step: Enhancing In-Context
  Learning Ability with Curriculum Learning}.
\newblock \bibinfo{journal}{\emph{arXiv preprint arXiv:2402.10738}}
  (\bibinfo{year}{2024}).
\newblock


\bibitem[Liu et~al\mbox{.}(2019)]%
        {liu2019roberta}
\bibfield{author}{\bibinfo{person}{Yinhan Liu}, \bibinfo{person}{Myle Ott},
  \bibinfo{person}{Naman Goyal}, \bibinfo{person}{Jingfei Du},
  \bibinfo{person}{Mandar Joshi}, \bibinfo{person}{Danqi Chen},
  \bibinfo{person}{Omer Levy}, \bibinfo{person}{Mike Lewis},
  \bibinfo{person}{Luke Zettlemoyer}, {and} \bibinfo{person}{Veselin
  Stoyanov}.} \bibinfo{year}{2019}\natexlab{}.
\newblock \showarticletitle{Roberta: A robustly optimized bert pretraining
  approach}.
\newblock \bibinfo{journal}{\emph{arXiv preprint arXiv:1907.11692}}
  (\bibinfo{year}{2019}).
\newblock


\bibitem[Liu et~al\mbox{.}(2013)]%
        {liu2013abstract}
\bibfield{author}{\bibinfo{person}{Yuanchao Liu}, \bibinfo{person}{Feng Wu},
  \bibinfo{person}{Ming Liu}, {and} \bibinfo{person}{Bingquan Liu}.}
  \bibinfo{year}{2013}\natexlab{}.
\newblock \showarticletitle{Abstract sentence classification for scientific
  papers based on transductive SVM}.
\newblock \bibinfo{journal}{\emph{Computer and Information Science}}
  \bibinfo{volume}{6}, \bibinfo{number}{4} (\bibinfo{year}{2013}),
  \bibinfo{pages}{125}.
\newblock


\bibitem[Loshchilov and Hutter(2018)]%
        {loshchilov2018decoupled}
\bibfield{author}{\bibinfo{person}{Ilya Loshchilov} {and}
  \bibinfo{person}{Frank Hutter}.} \bibinfo{year}{2018}\natexlab{}.
\newblock \showarticletitle{Decoupled Weight Decay Regularization}. In
  \bibinfo{booktitle}{\emph{International Conference on Learning
  Representations}}.
\newblock


\bibitem[Lu et~al\mbox{.}(2018)]%
        {lu2018functional}
\bibfield{author}{\bibinfo{person}{Wei Lu}, \bibinfo{person}{Yong Huang},
  \bibinfo{person}{Yi Bu}, {and} \bibinfo{person}{Qikai Cheng}.}
  \bibinfo{year}{2018}\natexlab{}.
\newblock \showarticletitle{Functional structure identification of scientific
  documents in computer science}.
\newblock \bibinfo{journal}{\emph{Scientometrics}} \bibinfo{volume}{115},
  \bibinfo{number}{1} (\bibinfo{year}{2018}), \bibinfo{pages}{463--486}.
\newblock


\bibitem[Lu et~al\mbox{.}(2020)]%
        {wei2020recognition}
\bibfield{author}{\bibinfo{person}{Wei Lu}, \bibinfo{person}{Pengcheng Li},
  \bibinfo{person}{Guobiao Zhang}, {and} \bibinfo{person}{Qikai Cheng}.}
  \bibinfo{year}{2020}\natexlab{}.
\newblock \showarticletitle{Recognition of Lexical Functions in Academic Texts:
  Automatic Classification of Keywords Based on BERT Vectorization}.
\newblock \bibinfo{journal}{\emph{Journal of the China Society for Scientific
  and Technical Information}} \bibinfo{volume}{39}, \bibinfo{number}{12}
  (\bibinfo{year}{2020}), \bibinfo{pages}{1320--1329}.
\newblock


\bibitem[Lui(2012)]%
        {lui2012feature}
\bibfield{author}{\bibinfo{person}{Marco Lui}.}
  \bibinfo{year}{2012}\natexlab{}.
\newblock \showarticletitle{Feature stacking for sentence classification in
  evidence-based medicine}. In \bibinfo{booktitle}{\emph{Proceedings of the
  Australasian Language Technology Association Workshop 2012}}.
  \bibinfo{pages}{134--138}.
\newblock


\bibitem[Luo et~al\mbox{.}(2022)]%
        {luo2022combination}
\bibfield{author}{\bibinfo{person}{Zhuoran Luo}, \bibinfo{person}{Wei Lu},
  \bibinfo{person}{Jiangen He}, {and} \bibinfo{person}{Yuqi Wang}.}
  \bibinfo{year}{2022}\natexlab{}.
\newblock \showarticletitle{Combination of research questions and methods: A
  new measurement of scientific novelty}.
\newblock \bibinfo{journal}{\emph{Journal of Informetrics}}
  \bibinfo{volume}{16}, \bibinfo{number}{2} (\bibinfo{year}{2022}),
  \bibinfo{pages}{101282}.
\newblock


\bibitem[Nanba et~al\mbox{.}(2010)]%
        {nanba2010automatic}
\bibfield{author}{\bibinfo{person}{Hidetsugu Nanba}, \bibinfo{person}{Tomoki
  Kondo}, {and} \bibinfo{person}{Toshiyuki Takezawa}.}
  \bibinfo{year}{2010}\natexlab{}.
\newblock \showarticletitle{Automatic creation of a technical trend map from
  research papers and patents}. In \bibinfo{booktitle}{\emph{Proceedings of the
  3rd international workshop on Patent information retrieval}}.
  \bibinfo{pages}{11--16}.
\newblock


\bibitem[Paszke et~al\mbox{.}(2019)]%
        {paszke2019pytorch}
\bibfield{author}{\bibinfo{person}{Adam Paszke}, \bibinfo{person}{Sam Gross},
  \bibinfo{person}{Francisco Massa}, \bibinfo{person}{Adam Lerer},
  \bibinfo{person}{James Bradbury}, \bibinfo{person}{Gregory Chanan},
  \bibinfo{person}{Trevor Killeen}, \bibinfo{person}{Zeming Lin},
  \bibinfo{person}{Natalia Gimelshein}, \bibinfo{person}{Luca Antiga},
  {et~al\mbox{.}}} \bibinfo{year}{2019}\natexlab{}.
\newblock \showarticletitle{Pytorch: An imperative style, high-performance deep
  learning library}.
\newblock \bibinfo{journal}{\emph{Advances in neural information processing
  systems}}  \bibinfo{volume}{32} (\bibinfo{year}{2019}).
\newblock


\bibitem[Petroni et~al\mbox{.}(2019)]%
        {petroni2019language}
\bibfield{author}{\bibinfo{person}{Fabio Petroni}, \bibinfo{person}{Tim
  Rockt{\"a}schel}, \bibinfo{person}{Sebastian Riedel},
  \bibinfo{person}{Patrick Lewis}, \bibinfo{person}{Anton Bakhtin},
  \bibinfo{person}{Yuxiang Wu}, {and} \bibinfo{person}{Alexander Miller}.}
  \bibinfo{year}{2019}\natexlab{}.
\newblock \showarticletitle{Language Models as Knowledge Bases?}. In
  \bibinfo{booktitle}{\emph{Proceedings of the 2019 Conference on Empirical
  Methods in Natural Language Processing and the 9th International Joint
  Conference on Natural Language Processing (EMNLP-IJCNLP)}}.
  \bibinfo{pages}{2463--2473}.
\newblock


\bibitem[Pride and Knoth(2020)]%
        {pride2020authoritative}
\bibfield{author}{\bibinfo{person}{David Pride} {and} \bibinfo{person}{Petr
  Knoth}.} \bibinfo{year}{2020}\natexlab{}.
\newblock \showarticletitle{An authoritative approach to citation
  classification}. In \bibinfo{booktitle}{\emph{Proceedings of the ACM/IEEE
  Joint Conference on Digital Libraries in 2020}}. \bibinfo{pages}{337--340}.
\newblock


\bibitem[Qin and Zhang(2020)]%
        {qin2020using}
\bibfield{author}{\bibinfo{person}{Chenglei Qin} {and}
  \bibinfo{person}{Chengzhi Zhang}.} \bibinfo{year}{2020}\natexlab{}.
\newblock \showarticletitle{Using Hierarchical Attention Network Model to
  Recognize Structure Functions of Academic Articles}.
\newblock \bibinfo{journal}{\emph{Data Analysis and Knowledge Discovery}}
  (\bibinfo{year}{2020}), \bibinfo{pages}{1}.
\newblock


\bibitem[Qin and Zhang(2022)]%
        {qin2022structure}
\bibfield{author}{\bibinfo{person}{Chenglei Qin} {and}
  \bibinfo{person}{Chengzhi Zhang}.} \bibinfo{year}{2022}\natexlab{}.
\newblock \showarticletitle{Which structure of academic articles do referees
  pay more attention to?: perspective of peer review and full-text of academic
  articles}.
\newblock \bibinfo{journal}{\emph{Aslib Journal of Information Management}}
  \bibinfo{number}{ahead-of-print} (\bibinfo{year}{2022}).
\newblock


\bibitem[Qin and Eisner(2021)]%
        {qin2021learning}
\bibfield{author}{\bibinfo{person}{Guanghui Qin} {and} \bibinfo{person}{Jason
  Eisner}.} \bibinfo{year}{2021}\natexlab{}.
\newblock \showarticletitle{Learning How to Ask: Querying LMs with Mixtures of
  Soft Prompts}. In \bibinfo{booktitle}{\emph{Proceedings of the 2021
  Conference of the North American Chapter of the Association for Computational
  Linguistics: Human Language Technologies}}. \bibinfo{pages}{5203--5212}.
\newblock


\bibitem[Qiu et~al\mbox{.}(2020)]%
        {qiu2020pre}
\bibfield{author}{\bibinfo{person}{Xipeng Qiu}, \bibinfo{person}{Tianxiang
  Sun}, \bibinfo{person}{Yige Xu}, \bibinfo{person}{Yunfan Shao},
  \bibinfo{person}{Ning Dai}, {and} \bibinfo{person}{Xuanjing Huang}.}
  \bibinfo{year}{2020}\natexlab{}.
\newblock \showarticletitle{Pre-trained models for natural language processing:
  A survey}.
\newblock \bibinfo{journal}{\emph{Science China Technological Sciences}}
  \bibinfo{volume}{63}, \bibinfo{number}{10} (\bibinfo{year}{2020}),
  \bibinfo{pages}{1872--1897}.
\newblock


\bibitem[Radford et~al\mbox{.}(2018)]%
        {radford2018improving}
\bibfield{author}{\bibinfo{person}{Alec Radford}, \bibinfo{person}{Karthik
  Narasimhan}, \bibinfo{person}{Tim Salimans}, {and} \bibinfo{person}{Ilya
  Sutskever}.} \bibinfo{year}{2018}\natexlab{}.
\newblock \showarticletitle{Improving language understanding by generative
  pre-training}.
\newblock  (\bibinfo{year}{2018}).
\newblock


\bibitem[Radford et~al\mbox{.}(2019)]%
        {radford2019language}
\bibfield{author}{\bibinfo{person}{Alec Radford}, \bibinfo{person}{Jeffrey Wu},
  \bibinfo{person}{Rewon Child}, \bibinfo{person}{David Luan},
  \bibinfo{person}{Dario Amodei}, \bibinfo{person}{Ilya Sutskever},
  {et~al\mbox{.}}} \bibinfo{year}{2019}\natexlab{}.
\newblock \showarticletitle{Language models are unsupervised multitask
  learners}.
\newblock \bibinfo{journal}{\emph{OpenAI blog}} \bibinfo{volume}{1},
  \bibinfo{number}{8} (\bibinfo{year}{2019}), \bibinfo{pages}{9}.
\newblock


\bibitem[Ruch et~al\mbox{.}(2007)]%
        {ruch2007using}
\bibfield{author}{\bibinfo{person}{Patrick Ruch}, \bibinfo{person}{Celia
  Boyer}, \bibinfo{person}{Christine Chichester}, \bibinfo{person}{Imad
  Tbahriti}, \bibinfo{person}{Antoine Geissb{\"u}hler}, \bibinfo{person}{Paul
  Fabry}, \bibinfo{person}{Julien Gobeill}, \bibinfo{person}{Violaine Pillet},
  \bibinfo{person}{Dietrich Rebholz-Schuhmann}, \bibinfo{person}{Christian
  Lovis}, {et~al\mbox{.}}} \bibinfo{year}{2007}\natexlab{}.
\newblock \showarticletitle{Using argumentation to extract key sentences from
  biomedical abstracts}.
\newblock \bibinfo{journal}{\emph{International journal of medical
  informatics}} \bibinfo{volume}{76}, \bibinfo{number}{2-3}
  (\bibinfo{year}{2007}), \bibinfo{pages}{195--200}.
\newblock


\bibitem[Schick et~al\mbox{.}(2020)]%
        {schick2020automatically}
\bibfield{author}{\bibinfo{person}{Timo Schick}, \bibinfo{person}{Helmut
  Schmid}, {and} \bibinfo{person}{Hinrich Sch{\"u}tze}.}
  \bibinfo{year}{2020}\natexlab{}.
\newblock \showarticletitle{Automatically identifying words that can serve as
  labels for few-shot text classification}.
\newblock \bibinfo{journal}{\emph{arXiv preprint arXiv:2010.13641}}
  (\bibinfo{year}{2020}).
\newblock


\bibitem[Schick and Sch{\"u}tze(2021)]%
        {schick2021exploiting}
\bibfield{author}{\bibinfo{person}{Timo Schick} {and} \bibinfo{person}{Hinrich
  Sch{\"u}tze}.} \bibinfo{year}{2021}\natexlab{}.
\newblock \showarticletitle{Exploiting Cloze-Questions for Few-Shot Text
  Classification and Natural Language Inference}. In
  \bibinfo{booktitle}{\emph{Proceedings of the 16th Conference of the European
  Chapter of the Association for Computational Linguistics: Main Volume}}.
  \bibinfo{pages}{255--269}.
\newblock


\bibitem[Shin et~al\mbox{.}(2020)]%
        {shin2020autoprompt}
\bibfield{author}{\bibinfo{person}{Taylor Shin}, \bibinfo{person}{Yasaman
  Razeghi}, \bibinfo{person}{Robert~L Logan~IV}, \bibinfo{person}{Eric
  Wallace}, {and} \bibinfo{person}{Sameer Singh}.}
  \bibinfo{year}{2020}\natexlab{}.
\newblock \showarticletitle{AutoPrompt: Eliciting Knowledge from Language
  Models with Automatically Generated Prompts}. In
  \bibinfo{booktitle}{\emph{Proceedings of the 2020 Conference on Empirical
  Methods in Natural Language Processing (EMNLP)}}.
  \bibinfo{pages}{4222--4235}.
\newblock


\bibitem[Sun et~al\mbox{.}(2021)]%
        {sun2021meda}
\bibfield{author}{\bibinfo{person}{Pengfei Sun}, \bibinfo{person}{Yawen
  Ouyang}, \bibinfo{person}{Wenming Zhang}, {and} \bibinfo{person}{Xin-yu
  Dai}.} \bibinfo{year}{2021}\natexlab{}.
\newblock \showarticletitle{{MEDA: Meta-Learning with Data Augmentation for
  Few-Shot Text Classification}}. In \bibinfo{booktitle}{\emph{Proceedings of
  the Thirtieth International Joint Conference on Artificial Intelligence,
  {IJCAI-21}}}, \bibfield{editor}{\bibinfo{person}{Zhi-Hua Zhou}} (Ed.).
  \bibinfo{publisher}{International Joint Conferences on Artificial
  Intelligence Organization}, \bibinfo{pages}{3929--3935}.
\newblock


\bibitem[Sung et~al\mbox{.}(2019)]%
        {sung2019pre}
\bibfield{author}{\bibinfo{person}{Chul Sung}, \bibinfo{person}{Tejas
  Dhamecha}, \bibinfo{person}{Swarnadeep Saha}, \bibinfo{person}{Tengfei Ma},
  \bibinfo{person}{Vinay Reddy}, {and} \bibinfo{person}{Rishi Arora}.}
  \bibinfo{year}{2019}\natexlab{}.
\newblock \showarticletitle{Pre-training BERT on domain resources for short
  answer grading}. In \bibinfo{booktitle}{\emph{Proceedings of the 2019
  Conference on Empirical Methods in Natural Language Processing and the 9th
  International Joint Conference on Natural Language Processing
  (EMNLP-IJCNLP)}}. \bibinfo{pages}{6071--6075}.
\newblock


\bibitem[Teufel et~al\mbox{.}(2006)]%
        {teufel2006automatic}
\bibfield{author}{\bibinfo{person}{Simone Teufel}, \bibinfo{person}{Advaith
  Siddharthan}, {and} \bibinfo{person}{Dan Tidhar}.}
  \bibinfo{year}{2006}\natexlab{}.
\newblock \showarticletitle{Automatic classification of citation function}. In
  \bibinfo{booktitle}{\emph{Proceedings of the 2006 conference on empirical
  methods in natural language processing}}. \bibinfo{pages}{103--110}.
\newblock


\bibitem[Wang et~al\mbox{.}(2019)]%
        {jiamin2019research}
\bibfield{author}{\bibinfo{person}{Jiamin Wang}, \bibinfo{person}{Wei Lu},
  \bibinfo{person}{Jiawei Liu}, {and} \bibinfo{person}{Qikai Cheng}.}
  \bibinfo{year}{2019}\natexlab{}.
\newblock \showarticletitle{Research on structure function recognition of
  academic text based on multi-level fusion}.
\newblock \bibinfo{journal}{\emph{Library and Information Service}}
  \bibinfo{volume}{63}, \bibinfo{number}{13} (\bibinfo{year}{2019}),
  \bibinfo{pages}{95}.
\newblock


\bibitem[Wang et~al\mbox{.}(2020)]%
        {qian2020structure}
\bibfield{author}{\bibinfo{person}{Qian Wang}, \bibinfo{person}{Jin Zeng},
  \bibinfo{person}{Jiawei Liu}, {and} \bibinfo{person}{Yue Qi}.}
  \bibinfo{year}{2020}\natexlab{}.
\newblock \showarticletitle{Structure Function Recognition of Academic Text
  Paragraph Based on Deep Learning}.
\newblock \bibinfo{journal}{\emph{Information Science (In Chinese)}}
  \bibinfo{volume}{38}, \bibinfo{number}{03} (\bibinfo{year}{2020}),
  \bibinfo{pages}{64--69}.
\newblock


\bibitem[Wei and Zou(2019)]%
        {wei2019eda}
\bibfield{author}{\bibinfo{person}{Jason Wei} {and} \bibinfo{person}{Kai Zou}.}
  \bibinfo{year}{2019}\natexlab{}.
\newblock \showarticletitle{EDA: Easy Data Augmentation Techniques for Boosting
  Performance on Text Classification Tasks}. In
  \bibinfo{booktitle}{\emph{Proceedings of the 2019 Conference on Empirical
  Methods in Natural Language Processing and the 9th International Joint
  Conference on Natural Language Processing (EMNLP-IJCNLP)}}.
  \bibinfo{pages}{6382--6388}.
\newblock


\bibitem[Wolf et~al\mbox{.}(2020)]%
        {wolf2020transformers}
\bibfield{author}{\bibinfo{person}{Thomas Wolf}, \bibinfo{person}{Lysandre
  Debut}, \bibinfo{person}{Victor Sanh}, \bibinfo{person}{Julien Chaumond},
  \bibinfo{person}{Clement Delangue}, \bibinfo{person}{Anthony Moi},
  \bibinfo{person}{Pierric Cistac}, \bibinfo{person}{Tim Rault},
  \bibinfo{person}{R{\'e}mi Louf}, \bibinfo{person}{Morgan Funtowicz},
  {et~al\mbox{.}}} \bibinfo{year}{2020}\natexlab{}.
\newblock \showarticletitle{Transformers: State-of-the-art natural language
  processing}. In \bibinfo{booktitle}{\emph{Proceedings of the 2020 conference
  on empirical methods in natural language processing: system demonstrations}}.
  \bibinfo{pages}{38--45}.
\newblock


\bibitem[Xie et~al\mbox{.}(2020)]%
        {xie2020unsupervised}
\bibfield{author}{\bibinfo{person}{Qizhe Xie}, \bibinfo{person}{Zihang Dai},
  \bibinfo{person}{Eduard Hovy}, \bibinfo{person}{Thang Luong}, {and}
  \bibinfo{person}{Quoc Le}.} \bibinfo{year}{2020}\natexlab{}.
\newblock \showarticletitle{Unsupervised data augmentation for consistency
  training}.
\newblock \bibinfo{journal}{\emph{Advances in Neural Information Processing
  Systems}}  \bibinfo{volume}{33} (\bibinfo{year}{2020}),
  \bibinfo{pages}{6256--6268}.
\newblock


\bibitem[Xu and Zhang(2024)]%
        {xu2024misconfidence}
\bibfield{author}{\bibinfo{person}{Shangqing Xu} {and} \bibinfo{person}{Chao
  Zhang}.} \bibinfo{year}{2024}\natexlab{}.
\newblock \showarticletitle{Misconfidence-based demonstration selection for llm
  in-context learning}.
\newblock \bibinfo{journal}{\emph{arXiv preprint arXiv:2401.06301}}
  (\bibinfo{year}{2024}).
\newblock


\bibitem[Yao et~al\mbox{.}(2021)]%
        {yao2021knowledge}
\bibfield{author}{\bibinfo{person}{Huaxiu Yao}, \bibinfo{person}{Ying-xin Wu},
  \bibinfo{person}{Maruan Al-Shedivat}, {and} \bibinfo{person}{Eric Xing}.}
  \bibinfo{year}{2021}\natexlab{}.
\newblock \showarticletitle{Knowledge-Aware Meta-learning for Low-Resource Text
  Classification}. In \bibinfo{booktitle}{\emph{Proceedings of the 2021
  Conference on Empirical Methods in Natural Language Processing}}.
  \bibinfo{pages}{1814--1821}.
\newblock


\bibitem[Yenicelik et~al\mbox{.}(2020)]%
        {yenicelik2020does}
\bibfield{author}{\bibinfo{person}{David Yenicelik}, \bibinfo{person}{Florian
  Schmidt}, {and} \bibinfo{person}{Yannic Kilcher}.}
  \bibinfo{year}{2020}\natexlab{}.
\newblock \showarticletitle{How does BERT capture semantics? A closer look at
  polysemous words}. In \bibinfo{booktitle}{\emph{Proceedings of the Third
  BlackboxNLP Workshop on Analyzing and Interpreting Neural Networks for NLP}}.
  \bibinfo{pages}{156--162}.
\newblock


\bibitem[Yin et~al\mbox{.}(2021)]%
        {yin2021mrt}
\bibfield{author}{\bibinfo{person}{Da Yin}, \bibinfo{person}{Weng~Lam Tam},
  \bibinfo{person}{Ming Ding}, {and} \bibinfo{person}{Jie Tang}.}
  \bibinfo{year}{2021}\natexlab{}.
\newblock \showarticletitle{MRT: Tracing the Evolution of Scientific
  Publications}.
\newblock \bibinfo{journal}{\emph{IEEE Transactions on Knowledge and Data
  Engineering}} (\bibinfo{year}{2021}).
\newblock


\bibitem[Yin et~al\mbox{.}(2019)]%
        {yin2019benchmarking}
\bibfield{author}{\bibinfo{person}{Wenpeng Yin}, \bibinfo{person}{Jamaal Hay},
  {and} \bibinfo{person}{Dan Roth}.} \bibinfo{year}{2019}\natexlab{}.
\newblock \showarticletitle{Benchmarking Zero-shot Text Classification:
  Datasets, Evaluation and Entailment Approach}. In
  \bibinfo{booktitle}{\emph{Proceedings of the 2019 Conference on Empirical
  Methods in Natural Language Processing and the 9th International Joint
  Conference on Natural Language Processing (EMNLP-IJCNLP)}}.
  \bibinfo{pages}{3914--3923}.
\newblock


\bibitem[Yu et~al\mbox{.}(2020)]%
        {yu2020identifying}
\bibfield{author}{\bibinfo{person}{Wenhao Yu}, \bibinfo{person}{Mengxia Yu},
  \bibinfo{person}{Tong Zhao}, {and} \bibinfo{person}{Meng Jiang}.}
  \bibinfo{year}{2020}\natexlab{}.
\newblock \showarticletitle{Identifying referential intention with
  heterogeneous contexts}. In \bibinfo{booktitle}{\emph{Proceedings of The Web
  Conference 2020}}. \bibinfo{pages}{962--972}.
\newblock


\bibitem[Zhang et~al\mbox{.}(2021)]%
        {zhang2021multiword}
\bibfield{author}{\bibinfo{person}{Guobiao Zhang}, \bibinfo{person}{Pengcheng
  Li}, \bibinfo{person}{Wei Lu}, {and} \bibinfo{person}{Qikai Cheng}.}
  \bibinfo{year}{2021}\natexlab{}.
\newblock \showarticletitle{Research on Keyword Semantic Function Recognition
  Based on Multi-feature Fusion}.
\newblock \bibinfo{journal}{\emph{Library and Information Service}}
  \bibinfo{volume}{65}, \bibinfo{number}{9} (\bibinfo{year}{2021}),
  \bibinfo{pages}{89}.
\newblock


\bibitem[Zhang et~al\mbox{.}(2022a)]%
        {zhang2022contrastive}
\bibfield{author}{\bibinfo{person}{Xin Zhang}, \bibinfo{person}{Fei Cai},
  \bibinfo{person}{Xuejun Hu}, \bibinfo{person}{Jianming Zheng}, {and}
  \bibinfo{person}{Honghui Chen}.} \bibinfo{year}{2022}\natexlab{a}.
\newblock \showarticletitle{A Contrastive learning-based Task Adaptation model
  for few-shot intent recognition}.
\newblock \bibinfo{journal}{\emph{Information Processing \& Management}}
  \bibinfo{volume}{59}, \bibinfo{number}{3} (\bibinfo{year}{2022}),
  \bibinfo{pages}{102863}.
\newblock


\bibitem[Zhang et~al\mbox{.}(2022b)]%
        {zhang2022active}
\bibfield{author}{\bibinfo{person}{Yiming Zhang}, \bibinfo{person}{Shi Feng},
  {and} \bibinfo{person}{Chenhao Tan}.} \bibinfo{year}{2022}\natexlab{b}.
\newblock \showarticletitle{Active example selection for in-context learning}.
\newblock \bibinfo{journal}{\emph{arXiv preprint arXiv:2211.04486}}
  (\bibinfo{year}{2022}).
\newblock


\bibitem[Zhou et~al\mbox{.}(2022)]%
        {zhou2022flipda}
\bibfield{author}{\bibinfo{person}{Jing Zhou}, \bibinfo{person}{Yanan Zheng},
  \bibinfo{person}{Jie Tang}, \bibinfo{person}{Li Jian}, {and}
  \bibinfo{person}{Zhilin Yang}.} \bibinfo{year}{2022}\natexlab{}.
\newblock \showarticletitle{FlipDA: Effective and Robust Data Augmentation for
  Few-Shot Learning}. In \bibinfo{booktitle}{\emph{Proceedings of the 60th
  Annual Meeting of the Association for Computational Linguistics (Volume 1:
  Long Papers)}}. \bibinfo{pages}{8646--8665}.
\newblock


\bibitem[Zhou and Zhang(2020)]%
        {zhou2020evaluating}
\bibfield{author}{\bibinfo{person}{Qingqing Zhou} {and}
  \bibinfo{person}{Chengzhi Zhang}.} \bibinfo{year}{2020}\natexlab{}.
\newblock \showarticletitle{Evaluating wider impacts of books via fine-grained
  mining on citation literatures}.
\newblock \bibinfo{journal}{\emph{Scientometrics}} \bibinfo{volume}{125},
  \bibinfo{number}{3} (\bibinfo{year}{2020}), \bibinfo{pages}{1923--1948}.
\newblock


\end{thebibliography}
\bibliographystyle{bib-style}
\end{document}